\documentclass[10pt,twocolumn,letterpaper,capitalize]{article}

\usepackage[pagenumbers]{cvpr}

\usepackage{amsmath,amsfonts}
\usepackage{graphicx}
\usepackage{caption}
\usepackage{subcaption}
\usepackage{svg}
\usepackage{array}
\usepackage{stfloats}
\usepackage{printlen}
\usepackage{booktabs}
\usepackage{algpseudocode}
\usepackage{algorithm}
\usepackage{placeins}

\def\x{{\mathbf x}}
\def\L{{\cal L}}
\def\E{{\mathbb{E}}}

\usepackage{xcolor}
\definecolor{cvprblue}{rgb}{0.21,0.49,0.74}
\usepackage[pagebackref,breaklinks,colorlinks,allcolors=cvprblue]{hyperref}

\def\paperID{12}
\def\confName{MORSE}
\def\confYear{2025}

\definecolor{cvprblue}{rgb}{0.21,0.49,0.74}
\usepackage[pagebackref,breaklinks,colorlinks,allcolors=cvprblue]{hyperref}

\def\paperID{12} 
\def\confName{MORSE}
\def\confYear{2025}

\title{A Sensor Agnostic Domain Generalization Framework for Leveraging Geospatial Foundation Models: Enhancing Semantic Segmentation via Synergistic Pseudo-Labeling and Generative Learning}

\author{
Anan Yaghmour\\
University of Houston\\
{\tt\small ayaghmou@cougarnet.uh.edu}
\and
Melba M. Crawford\\
Purdue University\\
{\tt\small mcrawford@purdue.edu}
\and
Saurabh Prasad\\
University of Houston\\
{\tt\small saurabh.prasad@ieee.org}
}

\begin{document}
\maketitle
\begin{abstract}
{Remote sensing enables a wide range of critical applications such as land cover, land use mapping, crop yield prediction, and environmental monitoring. Advances in satellite technology have expanded RS datasets, yet high-performance segmentation models remain dependent on extensive labeled data-challenged by annotation scarcity and variability across sensors, illumination, and geography. Domain adaptation offers a promising solution to improve model generalization. This paper introduces a domain generalization approach to leveraging emerging geospatial foundation models, by combining soft-alignment pseudo-labeling with source-to-target generative pre-training. We further provide new mathematical insights of MAE-based generative learning for domain-invariant feature learning. Experiments with hyperspectral and multispectral remote sensing datasets confirm our method’s effectiveness in enhancing adaptability and segmentation performance \footnote{Code available at: \url{https://github.com/anan0110692/GeoAI-FoundGen}}.}
\end{abstract}

\section{Introduction}
 Remote sensing (RS) is critical for addressing a range of environmental and societal challenges \cite{CVPR2024}. Semantic segmentation, which assigns a class label to each pixel in imagery data, has contributed to numerous fields, including RS. These include, but are not limited to, poverty estimation \cite{Jean2016}, crop yield prediction \cite{You2017}, and the detection of industrial pollution \cite{Lee2021}. Recent progress in satellite technology, including new satellite mission designs, high-resolution sensors, and more frequent revisit cycles, has significantly expanded the quantity  of available RS datasets \cite{Ref2}. For example, government programs such as Landsat \cite{CVPR24_REF_44} and Copernicus \cite{CVPR24_REF_11} produce large volumes of high-quality data that are publicly accessible \cite{CVPR2024}.

Deep learning has significantly advanced semantic segmentation in RS by offering strong capabilities in both feature extraction and classification \cite{Ref4}. However, realizing its full potential relies on access to large, labeled datasets, is extremely challenging. Creating pixel-level annotations is not only time-consuming but also labor-intensive. This difficulty is exemplified by the vast amount of unlabeled RS data and the shortage of high-quality annotated datasets, which are essential for advanced geospatial image analysis.

The sources of RS  provide heterogeneous data from different sensing modalities and the associated scales of data, as well as environmental conditions \cite{GU_REF_6}, \cite{CVPR2024}.
This variability limits the reusability of models, as their performance is hindered by disparities in datasets caused by these factors \cite{GU_REF_6}.

Domain adaptation (DA) can improve model generalization for geospatial image analysis, particularly when the available domain-specific dataset is limited. The core challenge in DA is managing the distribution shift between a well-labeled source domain and a different, but related, target domain \cite{Explora}. In the field of DA, various research efforts have approached this problem from different perspectives \cite{Explora}. Early work focused on traditional methods aimed at distribution alignment, which can be broadly categorized as three main approaches: instance-based \cite{REF7_M1}, \cite{REF7_M2}, feature-based \cite{REF7_M3}, \cite{REF7_M4} and classifier-based \cite{REF7_M5}, \cite{Ref7}. While each of  these classical DA methods have been successful, their strong reliance on hand-crafted features limits their ability to fully leverage the data \cite{Ref7}. In contrast, deep learning methods overcome these limitations by utilizing deep neural networks for automated feature learning. These approaches can be broadly divided into two categories: 1) discrepancy-based methods, which align marginal or conditional distributions between the source and target domains at one or multiple levels using techniques like maximum mean discrepancies (MMD) or adversarial training \cite{ADVENT}; and 2) self-training with pseudo-labels, where predictions from an ensemble model or a prior model state serve as pseudo-labels for the unlabeled target domain \cite{ADVENT}. A notable example of self-training in DA is ADVENT \cite{ADVENT}, which employs entropy minimization for soft-alignment of pseudo-labels over the target domain.

Influenced by the success of unsupervised pre-training in NLP \cite{FOUNDATION_LANG} and computer vision \cite{Foundation_CV}, which uses large unlabeled datasets to extract representations for downstream tasks \cite{PRO_REF_26},  \cite{Generative_pixel}, RS is now seeing a rise in foundational models. These models are pre-trained in a task-agnostic manner on vast unlabeled datasets using self-supervision and are later fine-tuned for specific applications \cite{PRO_REF_51}, \cite{PRO_REF_42}.

Self-supervised pre-training approaches come in two main variants: 1) Contrastive learning, which attracts similar samples and repels dissimilar ones \cite{PRO_REF_75}, and 2) Generative learning, which models the data distribution $P(X)$ to generate or reconstruct data from partial observations \cite{Generative_pixel}, \cite{PRO_REF_28}. Generative models fall into two main categories: auto-regressive (AR) models and auto-encoding (AE) models, such as Denoising Autoencoders (DAEs) \cite{RingMo}. The Masked Autoencoder (MAE), a type of DAE \cite{PRO_REF_26}, \cite{RingMo}, uses an encoder to map inputs to a latent space and a decoder to reconstruct them from masked patches \cite{PRO_REF_26}, \cite{PRO_REF_51}. This approach, based on the Vision Transformer (ViT) model \cite{PRO_REF_15}, reconstructs images using visible patches and mask tokens, with discrepancies measured by Mean Squared Error (MSE) \cite{PRO_REF_51}.

The rise of foundational models in RS has greatly improved performance across various tasks. For example, the RingMo model \cite{RingMo} achieved state-of-the-art results on multiple datasets for tasks such as scene recognition and semantic segmentation \cite{PRO_REF_42}. The transformer-based Prithvi model \cite{PRO_REF_51} released by NASA and IBM, pre-trained on over 1TB of multispectral satellite imagery, outperformed a conditional Generative Adversarial Network (GAN) by $5.7\%$ in multi-temporal cloud imputation. Similarly, SpectralGPT \cite{PRO_REF_28}, trained on over one million spectral images from Sentinel-2, has shown significant advancements in tasks such as scene classification, semantic segmentation, and change detection.

In this study, we leverage the generalization capacity of foundational models like Prithvi, combined with MAE-based generative learning, for DA in RS. { Leveraging Prithvi's strong capability in reconstructing multispectral images and its effective generalization across diverse Earth observation downstream tasks-including multi-temporal cloud gap imputation and wildfire scar segmentation, among others-Prithvi extends the applicability of RS foundation models to previously unexplored tasks \cite{prthivi}. Additionally, to evaluate the effectiveness and practicality of our proposed method, we leverage the fact that Prithvi was originally pre-trained on U.S. data. Therefore, we assess our approach on datasets from other continents, which are entirely unseen by Prithvi. We fine-tune its encoder-pre-trained on large RS datasets-to extract domain-invariant features}. These features assist the decoder in reconstructing masked patches within the target domain, utilizing data from both the source and unmasked target datasets, a process we refer to as \emph{source-to-target MAE generative learning}. { Our framework further enhances the adaptability of  RS foundation models, such as Prithvi, allowing them to handle diverse data modalities, including RGB, multispectral, and hyperspectral imaging. However, it is important to distinguish this adaptability from domain adaptation , which is the primary focus of this study, where we have a well-labeled source domain and we try to adapt the model to the target domain, which is distinct yet related and has limited or no labels.}
 To ensure feature alignment between domains, we employ the pre-trained source model for soft pseudo-label generation in the target domain through entropy minimization, as suggested by \cite{ADVENT}, while concurrently fine-tuning with a limited set of labeled target samples. This approach aims to improve the model's generalization and adaptability across domains. Additionally, we present a rigorous mathematical analysis of the effects of MAE-based generative learning on segmentation tasks.

\noindent {The main contributions of this work are as follows:}
\begin{enumerate}
    \item We extend the feature extraction capabilities of foundational models by introducing a pre-training framework adaptable to various RS data modalities. To the best of our knowledge, this is the first framework to adapt foundational models pre-trained using MAE with multi-spectral imagery for reconstruction of hyperspectral images.
    
    \item We propose a novel approach that combines soft-alignment pseudo-labeling with source-to-target MAE generative pre-training for effective DA of these foundation models for RS tasks. A key aspect of our paradigm is that foundation models trained on a specific widely available sensor can be adapted to a variety of different downstream imaging sensors depending upon the need. 
    
    \item We provide a rigorous mathematical analysis of the effects of MAE-based generative learning on segmentation tasks. {To wit, we show mathematically how in the proposed approach, MAE leverages unlabeled data through a dynamic weighting mechanism that is closely tied to the segmentation model's confidence with the target domain data.}
\end{enumerate}

\section{Related Work}
\noindent \textbf{RS Image Generation and RS Foundation Models}
Previous research on RS image generation has primarily focused on text-to-image and image-to-image tasks \cite{HSI_SCARCITY_META}. One example of image-to-image generation is modality transfer \cite{META_42}, where the primary objective is to transform between multimodal data types, such as RGB, infrared, SAR, and HSI. However, these methods, often constrained by small and limited datasets, lack diversity in land cover types and resolutions. These limitations, among others, have driven the adoption of foundation models in RS, which can leverage large-scale data and be adapted to a wide range of downstream tasks.

Due to the limited availability of high-quality labeled HSI data and its high-dimensional characteristics, few methods have utilized HSI for generative-based self-learning \cite{HSI_SCARCITY_META}. Consequently, most pre-training phases for RS foundation models-where MAE-based pre-training is widely used \cite{HSI_SCARCITY_META}-primarily rely on multispectral data, as seen in models like Prithvi and RingMo \cite{prthivi, RingMo}. In contrast, our proposed framework extends MAE-based generative learning specifically for RS foundation models across various modalities, including HSI.

\noindent \textbf{Pre-training and Model Generalization Ability}

Pre-training has seen a resurgence across various fields, including computer vision and NLP \cite{Generalization_ability}, following its pivotal role in advancing state-of-the-art performance in many NLP tasks \cite{Generative_pixel}. This cross-domain success is supported by a growing number of empirical studies demonstrating its effectiveness \cite{Generalization_ability}. However, despite its recent achievements, a thorough understanding of the key factors that impact the generalization performance of models using pre-training for target tasks remains limited \cite{Generalization_ability}, especially in comparison to observed empirical results \cite{Generalization_ability}.

Under the framework where pre-training involves an unsupervised learning objective, the model's feature extractor minimizes loss over an unlabeled dataset, followed by a fine-tuning stage where the pre-trained semantic feature extractor is used to train a task-specific network (e.g. classifier) on a small labeled dataset. A line of research has provided an upper bound on the generalization ability of fine-tuned models, similar in concept to the bounding approach used in the Probably Approximately Correct (PAC) theory. Here, generalization refers to the difference between the true loss, calculated over the actual data distribution, and the empirical loss, measured on the sampled training data. For example, \cite{Generalization_ability} conducted a generalization analysis of multi-layer transformer models with residual blocks, demonstrating that a model’s generalization ability depends on four key factors: representation transferability, Rademacher complexity, domain heterogeneity, and the generalization of the pre-training task. \cite{Pretraining_Regularizes} interpreted pre-training as a form of regularization, showing that unsupervised pre-training aids in learning meaningful Tikhonov matrices, which enhance the stability of deep learning algorithms and lead to faster model generalization with respect to sample size.

Alternatively, another line of research focuses on the generative aspect of pre-training, modeling either the joint distribution $P(X, Y)$ (where $X$ refers to the data and $Y$ to their corresponding label) or the marginal distribution $P(X)$, and exploring its connection to discriminative learning, specifically $P(Y|X)$. This work is exemplified by \cite{Cambridge}, which analyzed the framework utilizing generative learning to enhance discriminative learning, modeling it as a joint-likelihood maximization problem. This formulation is equivalent to simultaneously minimizing both the pre-training loss and the task-specific loss, which in our case is semantic segmentation. One key insight from this work is that generative learning can leverage unlabeled data by making predictions about the class of each unlabeled data point and using these predictions directly to regulate the influence of each data point on the discriminative learning process. Building on this idea, we have investigated the impact of MAE-based generative learning on the segmentation task, showing that it regulates the effect of unlabeled pixels from the target domain based on an expectation closely tied to the segmentation model’s confidence in these pixels.

\noindent \textbf{  Self-supervised Learning for DA}
\cite{CVPR2024} introduces Scaled Low-Rank (SLR) adapters with a small number of parameters into a pre-trained foundation model, using MAE-based self-supervised learning on unlabeled target domain data. After pre-training on the target domain, the model is adapted to the target task (e.g., classification) through supervised fine-tuning with a few labeled samples. In contrast, our method differs in its use of MAE self-supervised learning, allowing the decoder to reconstruct masked patches in the target domain by leveraging data from both the source and target domains, thereby facilitating the extraction of domain-invariant features. \cite{ADVENT} introduced an efficient self-supervised DA framework for semantic segmentation by proposing the use of an entropy loss to penalize low-confidence predictions in the target domain. This approach acts as a soft-assignment form of pseudo-labeling. \cite{Kim2021} proposed a cross-domain self-supervised learning approach for adaptation using a limited number of source labels. \cite{PCS} proposed Prototypical Cross-domain Self-supervised Learning, a single-stage framework that combines representation learning and domain alignment using few-shot labeled source samples. This approach not only aligns cross-domain low-level features but also encodes and aligns semantic structures within a shared embedding space across domains.
\vspace{-6pt}

\section{Method}
\vspace{-3pt}

\begin{figure}[!h]
    \centering
   \includegraphics[width=1\linewidth]{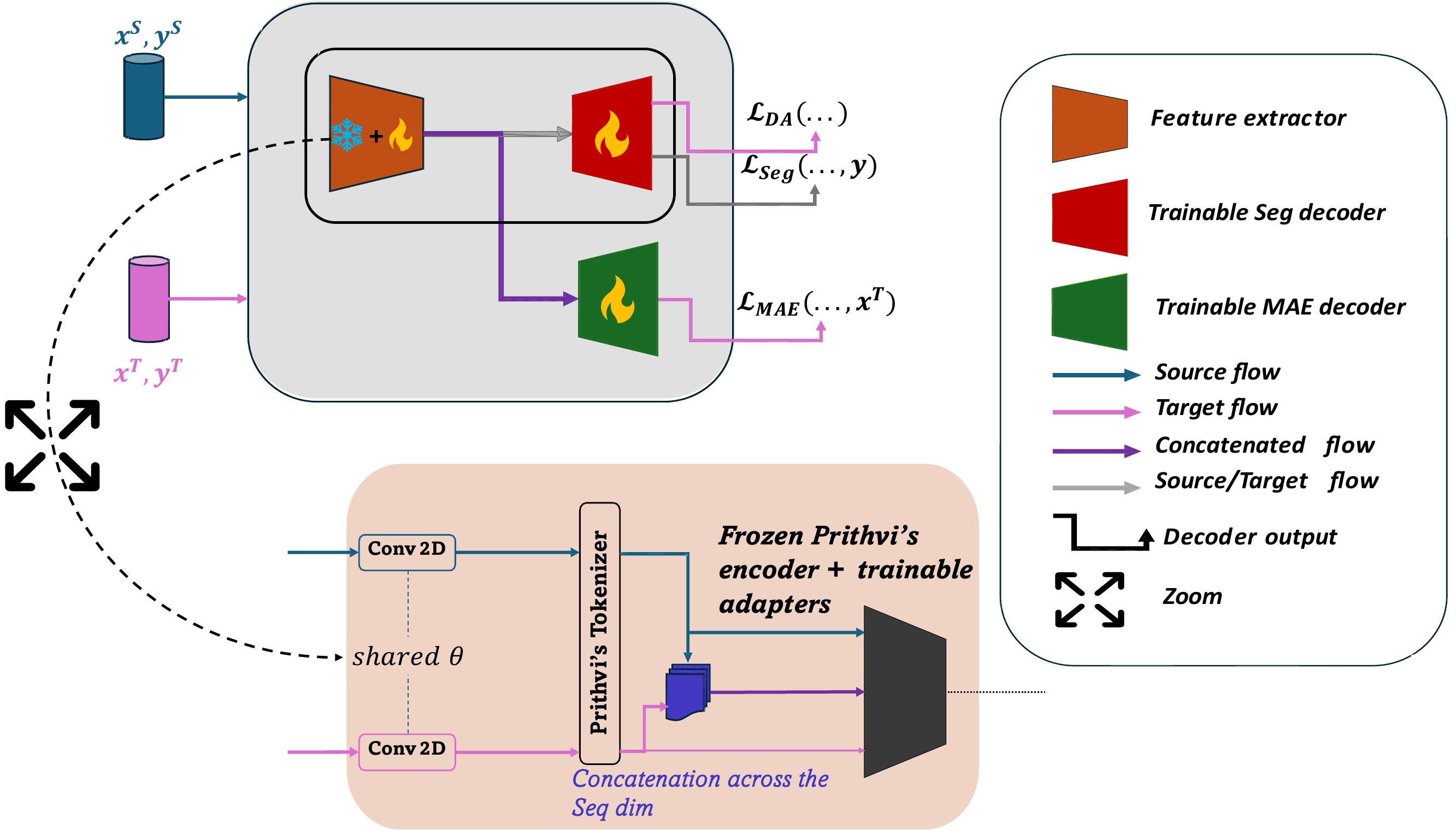}
    \caption{Illustration of the proposed multi-task learning framework, represented by a total loss function comprising three learning objectives, each with its associated loss term: the primary segmentation task (\(\mathcal{L}_{Seg}\)), and two auxiliary tasks-domain alignment (\(\mathcal{L}_{DA}\)) and MAE-based self-supervision (\(\mathcal{L}_{MAE}\)).}
    \label{fig:Method_diagram}
\end{figure}

\noindent  \textbf{Notations and Problem Statement:}
Let $\mathbf{S}=\{(\mathbf{X}^S_n, \mathbf{Y}^S_n) \mid \mathbf{X}^S_n =\{x_{n,k}^S \mid x_{n,k}^S \in \mathbb{R}^C\}_{k=1}^{H_S \times W_S},\, \mathbf{Y}^S_n =\{y_{n,k}^S \mid y_{n,k}^S \in \{1,..,C_S'\}\}_{k=1}^{H_S \times W_S}\}_{n=1}^{N_S}$ represent the source domain, where $\mathbf{X}^S_n$ is the image tensor, $\mathbf{Y}^S_n$ is the ground truth mask tensor, $N_S$ is the number of samples, $C$ is the number of optical channels, $H_S$ and $W_S$ are the height and width in pixels, and $C_S'$ denotes the total number of classes. Also, let $\mathbf{T}$ be the target domain, where $\mathbf{T}= \{(\mathbf{X}^{T}_n, \mathbf{Y}^{T}_n) \mid \mathbf{X}^{T}_n =\{x_{n,k}^T \mid x_{n,k}^T \in \mathbb{R}^C\}_{k=1}^{H_T \times W_T},\, \mathbf{Y}^{T}_n =\{y_{n,k}^T \mid y_{n,k}^T \in \{1,..,C_T'\}\}_{k=1}^{L}\}_{n=1}^{N_{T}}$  with $L \ll H_T \times W_T$, representing the limited number of labeled pixels in the target domain. Our semantic segmentation framework includes a feature extractor \(f_\theta\)-a composite of Prithvi’s frozen and other layers as we will demonstrate later-and a segmentation decoder \(h_{\theta_{seg}}\). The decoder used in the  MAE task is denoted by \(g_{\theta_{M}}\). \cref{fig:Method_diagram} illustrates the proposed multi-task learning framework, represented by the following  total loss function
\begin{align}
    \mathcal{L}_{Tot} &=& \mathcal{L}_{Seg} + \lambda_{DA}\mathcal{L}_{DA}  
+\lambda_{MAE}\mathcal{L}_{MAE} \label{Min_total_loss} 
\end{align}


As shown in the diagram, the semantic segmentation loss, $\mathcal{L}_{Seg}$, 
 primarily depends on the source domain, where ground truth labels are available, along with a limited number of labeled samples from the target domain.  The MAE loss, $\mathcal{L}_{MAE}$ , depends on the source domain and unmasked patches from the target domain. $\mathcal{L}_{DA}$, the domain alignment loss term, inspired by the self-learning DA framework in \cite{ADVENT}, is designed to penalize low-confidence predictions in the target domain. Consequently, as shown in \cref{fig:Method_diagram}, this loss relies exclusively on the target domain.

The zoomed region in \cref{fig:Method_diagram} illustrates the high-level architecture of the feature extractor, highlighting two primary components of our approach: 1) Adapting the Prithvi foundation model from its original MAE-based self-learning task using a specific multispectral dataset to learning tasks involving semantic segmentation with different sensor specifications (e.g. different optical channels, more channels - e.g. hyperspectral imagery etc.). This adaptation is achieved through the integration of Conv2D layers and the adapter tuning technique \cite{PRO_REF_75}, which incorporates additional trainable modules into Prithvi’s frozen encoder, as shown in \cref{fig:Method_diagram}. 2) Leveraging both the source domain and masked patches from the target domain. We guide the adapted encoder (Prithvi+ with added adapters) to extract features exhibiting domain-invariance by concatenating the source domain embedding sequence, which represents all patches  from the source domain image, with the unmasked patches  from the target domain along the sequence dimension. This configuration not only facilitates the learning of generalizable and correlated features across the two domains through the self-attention mechanism in ViT, but it also provides our framework with the flexibility to handle domains with different spatial dimensions. This is because the concatenation occurs in the embedding space rather than in the data space, allowing for the integration of images from different domains without spatial constraints.

\subsection{Segmentation Task}
 The segmentation model, represented by $ \mathbb{P}(y_i \mid \mathbf{X}; \theta, \theta_{\text{seg}}) = h_{\theta_{\text{seg}}}(f_\theta(\mathbf{X}), y_i)$, estimates the probability of class $y$ at the $i^{\text{th}}$ pixel given the entire image $\mathbf{X}$. It is trained in a supervised manner on both $\mathbf{S}$ and $\mathbf{T}$, using available labels from each domain. This process involves solving the optimization problem:
\vspace{-6pt}
\begin{align}
\underset{\theta,\theta_{seg}}{\mathrm{argmin}} \, \mathcal{L}_{Seg} = 
\mathbb{E}_{(\mathbf{X}_n, \mathbf{Y}_n) \in \mathbf{S} \cup \mathbf{T}} 
\left[ \mathcal{L}_{MCE}\big(\mathbf{X}_n, \mathbf{Y}_n \big) \right] 
\end{align}

\noindent where the mean cross-entropy loss, averaged over all pixels, $\mathcal{L}_{MCE}\big(\mathbf{X}_n, \mathbf{Y}_n \big)$, is defined as:
\begin{align}
\mathcal{L}_{MCE}\big(\mathbf{X}_n, \mathbf{Y}_n \big) =  
\mathbb{E}_{y_{n,k} \in \mathbf{Y}_n} 
\Big[ & -\log h_{\theta_{seg}}(f_\theta(\mathbf{X}_n), y_{n,k}) \Big]
\end{align}

\subsection{MAE Task}
We express the target domain images as a union of masked and unmasked patches, $\mathbf{X}_n^T = \mathbf{X}_n^{T_m} \cup \mathbf{X}_n^{T_u}$, along with the generative model represented by $\hat{x}^T_{n,k} = g_{\theta_{M}}(f_\theta(\mathbf{X}_n^S, \mathbf{X}_n^{T_u}), k)$, which denotes the predicted channel vector at pixel $k$ of the $n^{\text{th}}$ image in the target domain. The MAE task can be learned through minimizing $\mathcal{L}_{MAE}$ as follows:
\begin{align}
\underset{\theta, \theta_{M}}{\mathrm{argmin}} \, \mathcal{L}_{MAE} = & \, 
\E_{(\mathbf{X}_n^T, \mathbf{X}_n^S) \in \mathbf{S} \cup \mathbf{T}} \nonumber \\
& \left[ \mathcal{L}_{MMSE}(\mathbf{X}_n^S, \mathbf{X}_n^T) \right]
\end{align}

\noindent   The mean MSE, averaged over all pixels, $\mathcal{L}_{MMSE}(\mathbf{X}_n^S, \mathbf{X}_n^T)$, is defined as follows:
\begin{align}
  \mathcal{L}_{MMSE}(\mathbf{X}_n^S, \mathbf{X}_n^T) = \nonumber \\
  \mathbb{E}_{x_{n,k}^{T_m} \in \mathbf{X}_n^{T_m}} \Bigg[ &\frac{\Big\| g\big(f(\mathbf{X}_n^S, \mathbf{X}_n^{T_u}), k\big) 
  - x_{n,k}^{T_m} \Big\|_2^2}{C} \Bigg] 
\end{align}

\subsection {Domain Alignment Task}
Motivated by soft pseudo-labeling based fine-tuning, we encourage our segmentation model to generate more confident predictions on the target domain by minimizing the loss term $\mathcal{L}_{DA}$ .

\begin{align}
 \mathcal{L}_{DA} =  \mathbb{E}_{\mathbf{X}_n^T \in T} \left[ \mathcal{L}_{ME}(\mathbf{X}_n^T) \right]
 \label{eqn:DA}
\end{align}

\noindent Here $\mathcal{L}_{ME}(X_n^T)$, the pixel-wise  mean Shanon Entropy is given as follows:
\begin{align}
 \mathcal{L}_{ME}(\mathbf{X}_n^T) &= \underset{k\in \operatorname{Index}(\mathbf{X}_n^{T})}{\mathbb{E}}
  \Bigg[ \frac{-1}{\log(C_T')} \sum_{y_{n,k} }h_{\theta_{\text{seg}}}(f_\theta(\mathbf{X}_n^T),y_{n,k}) \nonumber \\
 & \quad \times \log h_{\theta_{\text{seg}}}(f_\theta(\mathbf{X}_n^T), y_{n,k}) \Bigg] 
\end{align}

\subsection{Mathematical insights of the proposed approach}
Here, we provide a mathematical insight on the impact of MAE learning on the segmentation task in the proposed approach by setting $\lambda_{DA} = 0$ in \cref{Min_total_loss}. Hence
\vspace{-6pt}

\begin{align}
    \underset{\theta,\theta_{seg},\theta_{M}}{\mathrm{argmin}} \,\mathcal{L}_{Seg}+ \mathcal{L}_{MAE}&=& \underset{\theta,\theta_{seg},\theta_{M}}{\mathrm{argmax}}\, J(\theta,\theta_{seg},\theta_{M})
\end{align}
 $J$ represents the joint probability of the observed data, as defined in the supplementary material. To explore MAE's influence on segmentation learning when optimizing \( J \) through gradient-based methods, we compute \( \frac{\partial J}{\partial \theta} \) as \( \theta \) is shared between the two tasks, facilitating this interaction. For clarity, from $\frac{\partial J}{\partial \theta}$ expression, we focus solely on the MAE-related term for unlabeled target domain samples. While some notation is simplified here, a complete formal explanation and proof is provided in the supplementary material (Sec. A).
\vspace{-6pt}
\begin{align}
\frac{\partial J}{\partial \theta} &= 
\hdots  \nonumber \\
&\quad + \sum_{k\in \operatorname{Index}(\mathbf{X}_n^{T})\setminus \operatorname{Index}(\mathbf{Y}_n^{T}) } 
\sum_{y_{n,k}} \phi_1( h_{\theta_{seg}}(f_\theta(\mathbf{X}_n^T), y_{n,k})) \nonumber \\
&\qquad \times \frac{\partial}{\partial \theta} 
\log \phi_2( h_{\theta_{seg}}(f_\theta(\mathbf{X}_n^T), y_{n,k}))\label{eqn:impact}
\end{align}
Where $\phi_1(.)$ is the expected value defined in the supplementary material 
(Eq. (26) in Sec. A.2.2)
and $\phi_2(.)= \mathbb{P}(x_k^T, y_k \mid \mathbf{X}_S; \bar{\theta}) $ which is a function of $h_{\theta_{seg}}(f_\theta(\mathbf{X}_n^T), y_{n,k})$. 

From \cref{eqn:impact}, we observe how generative learning on unlabeled samples from the target domain influences segmentation learning . For each class \( y_{n,k} \), the expectation, denoted by \(\phi_1\), dynamically regulates the impact of each unlabeled pixel \( k \) on learning \( h_{\theta_{seg}}(f_\theta(\mathbf{X}_n^T), y_{n,k}) \), encoded by \(\phi_2\). This regulation depends on the model’s confidence in assigning pixel \( k \) to class \( y_{n,k} \); highly confident regions exert greater influence on segmentation learning, while less confident regions contribute less. Furthermore, incorporating \(\mathcal{L}_{DA}\) benefits the model by encouraging it to identify and leverage confident regions, thus enhancing their influence on segmentation learning.

\section{Experiments \& Results}
\subsection{Experiment Setting}
\noindent \textbf{Datasets}.

To demonstrate both the broader applicability and effectiveness of our proposed method, we conducted experiments on two datasets representing two different sensing regimes: a hyperspectral dataset, specifically the HSI dataset from C2Seg-AB \cite{PRO_REF_31}, and a multispectral dataset, FLAIR \cite{FLAIR}.

The C2Seg-AB dataset, a subset of the multimodal RS C2Seg dataset, comprises a diverse set of hyperspectral, multispectral, and synthetic aperture radar (SAR) data, as detailed in \cite{PRO_REF_31}. This subset captures urban scenes from Berlin and Augsburg, Germany, using data from the EnMAP, Sentinel-2, and Sentinel-1 satellite missions \cite{PRO_REF_31}. To maintain uniform spatial resolution across modalities, all images were upsampled to a 10-meter ground sampling distance (GSD) \cite{PRO_REF_31}. For this study, Berlin images were used at dimensions of 1000 × 500 pixels, while Augsburg images were used at 500 × 1000 pixels, both with 242 spectral bands ranging from 400 nm to 2500 nm. Berlin was designated as the source domain and Augsburg as the target domain, with both domains containing the same classes.

The FLAIR dataset includes very-high-resolution (VHR, 20 cm) images, photogrammetry-derived surface models, and Sentinel-2 multispectral satellite time series with approximately 5-day revisit intervals at the equator \cite{FLAIR}. It spans 817 km² across 50 French sub-regions characterized by diverse bioclimatic conditions, introducing complex domain shifts \cite{FLAIR}. The dataset has been collected from 916 unique locations within these domains, which are organized into similarly sized areas, further divided into uniform patches. For a visualization of the dataset architecture, see Fig. 2 in \cite{FLAIR_ARC}.

\begin{figure}
    \centering
    \includegraphics[width=.9\linewidth]{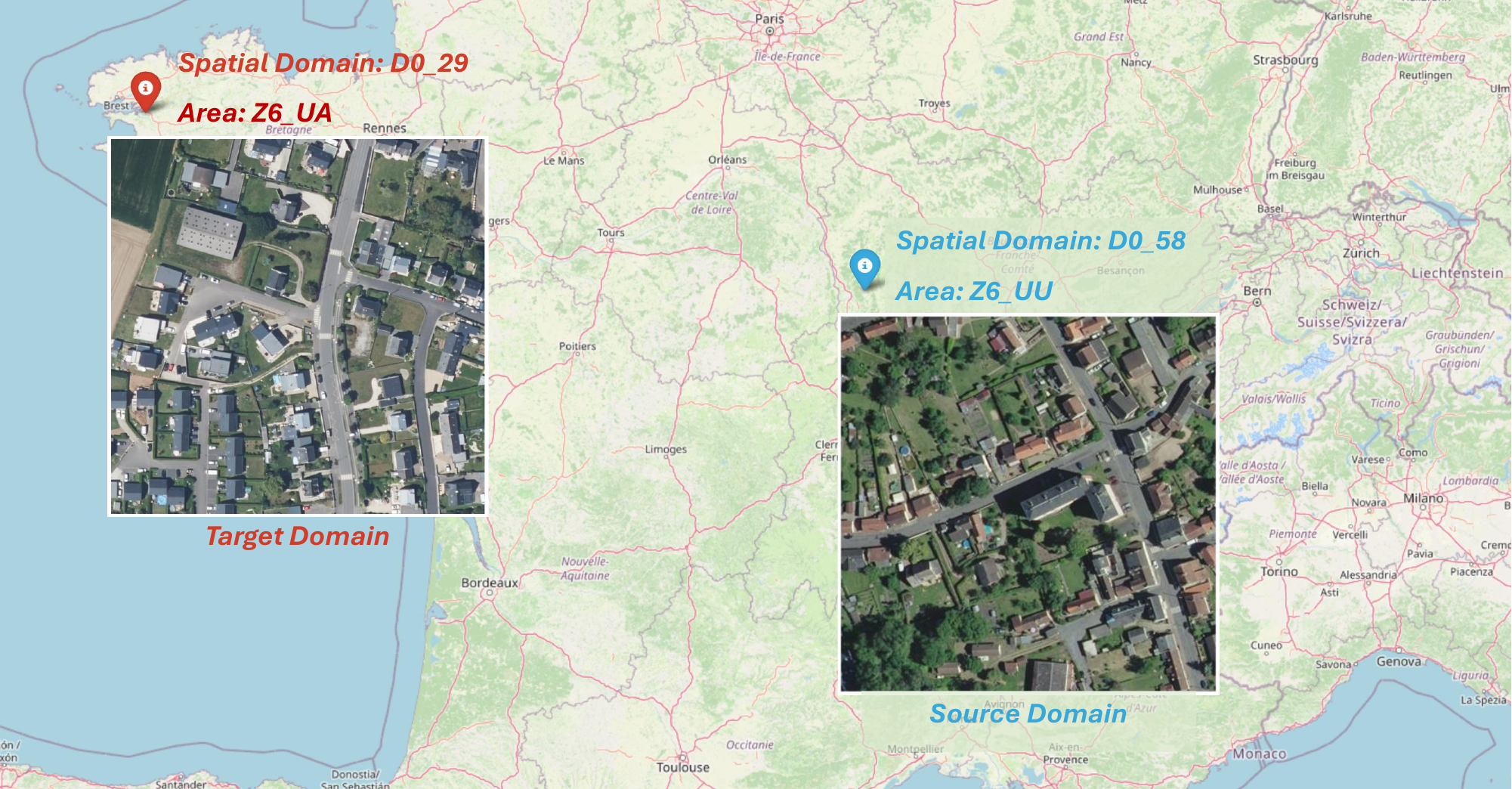}
    \caption{Illustration of the target and source domains used from the FLAIR dataset. The red and blue icons mark the geographical locations and data folders for the target and source domains, respectively. }
    \label{fig:enter-label}
\end{figure}

Leveraging the complex domain shifts presented by the FLAIR dataset, we consider a scenario where the target and source domains differ spatially, temporally, and in the number of classes. \cref{fig:enter-label} shows the target and source domains used in this study. The red icon marks the geographical location, spatial domain folder, and area subfolder from which the target domain data was obtained, while the blue icon provides the same details for the source domain. Additionally, patches from each area subfolder were combined into a single large tile, from which a 1000 × 1000 tile was selected for the pixel-level segmentation task, as displayed under each icon.

\noindent \textbf{Network Architecture}. 
\begin{figure*}
    \centering
    \includegraphics[width=.8\textwidth]{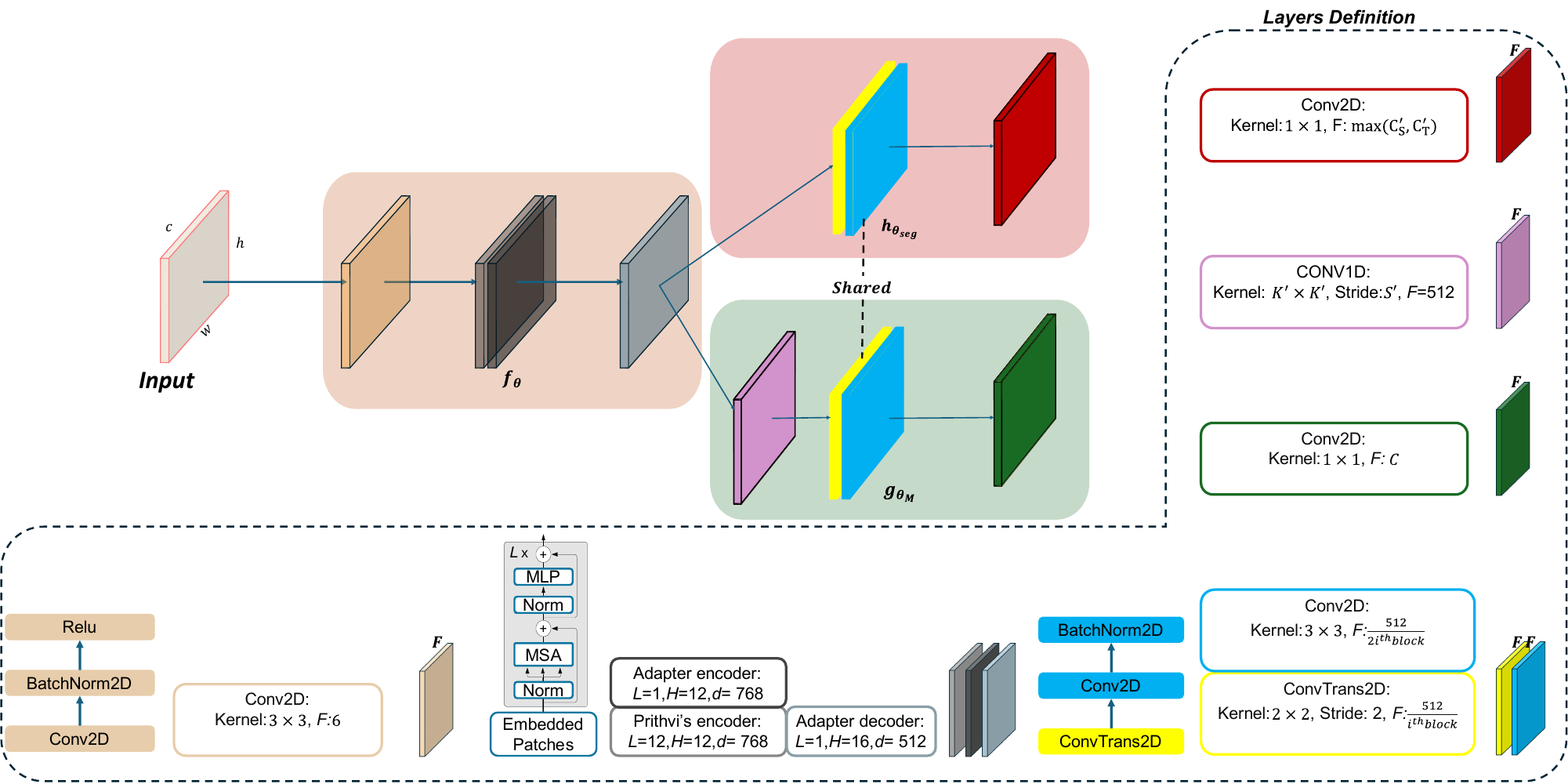}
    \caption{Prithvi model with adapters: feature extractor \(f_\theta\), segmentation head \(h_{\theta_{\text{Seg}}}\), and MAE generative head \(g_{\theta_M}\). Layers are color-coded. Parameters: \(L\) = transformer depth, \(H\) = attention heads, \(d\) = embedding dimension, \(F\) = output width; \(c\), \(h\), \(w\) = input channels, height, and width.
}
    \label{fig:Model}
\end{figure*}
\cref{fig:Model} illustrates the network architecture, consisting of three main components: the feature extractor $f_\theta$, the segmentation head $h_{\theta_{seg}}$, and the MAE-based generative head $g_{\theta_M}$. The feature extractor includes the Conv2D layer, Prithvi’s frozen encoder, and additional encoder-decoder adapters. The Conv2D layer is used for spectral adaptation, aligning Prithvi's spectral dimension of 6 with the spectral dimension of the target dataset. We employed an adapter tuning technique, which incorporates extra trainable modules into the original model \cite{PRO_REF_75}. Specifically, additional transformer layers-referred to as trainable encoder/decoder adapters-were added. During training, Prithvi’s original encoder remains frozen, while the added adapters are allowed to adjust, enabling the feature extractor to leverage Prithvi’s robust feature extraction while learning task-specific features with flexibility. 

To promote domain-invariance in the feature extractor and shared segmentation layers, we configure the generative flow, $g_{\theta_M} \circ f_{\theta}$, by concatenating the source domain embedding sequence, representing all source patches, with the unmasked target domain patches along the sequence dimension. This setup enables our framework to leverage ViT's self-attention for learning generalizable features across domains and provides flexibility for handling domains with different spatial dimensions. To recover the target domain sequence from the concatenated source-target sequence, it is passed through a Conv1D layer operating over the sequence dimension, with kernel size $K' \times K'$ and stride $S'$. Parameters $K'$ and $S'$ are chosen to ensure the correct target domain sequence length, $\left\lfloor \frac{H_T}{p} \right\rfloor \times \left\lfloor \frac{W_T}{p} \right\rfloor$, where $p=16$ is the patch size of the Prithvi model. This ``blending" effect of the Conv1D layer between source and target embeddings encourages the model to derive target-specific features from the source domain, particularly when the target image has a high masking ratio, leaving only a small number of unmasked patches.

Both heads, $h_{\theta_{seg}}$ and $g_{\theta_M}$, share four blocks of (ConvTranspose2D → Conv2D → BatchNorm2D) layers that act as upsamplers, restoring the input's original resolution and simultaneously learning features beneficial for both segmentation and generative tasks. Each head then applies a Conv2D layer with a 1x1 kernel and an appropriate number of filters ($F$) specific to its task. For the segmentation head, $F$ corresponds to the maximum number of classes across the two domains, allowing for open-set DA where one domain's classes may be a subset of the other's. For the generative head, $F$ matches the spectral dimension length of the data.
Additional implementation details can be found in the supplementary material.

\subsection{Results}
\noindent \textbf{C2Seg-AB Results}

\begin{table}[htbp]
  \centering
  \caption{C2Seg-AB Results-upper section shows the F1 scores by class, and the lower section details the mean and standard deviation for MA, mIoU, and mF1 metrics.}
  \resizebox{\columnwidth}{!}
  {%
   \begin{tabular}{lrrrrrrrr}
    \toprule
    \multicolumn{1}{c}{\textbf{Classes}} & \multicolumn{1}{c}{\textbf{Zero Shot}} & \multicolumn{1}{c}{\textbf{GDA \cite{CVPR2024}}} & \multicolumn{1}{c}{\textbf{PCS \cite{PCS}}} & \multicolumn{1}{c}{\textbf{CDS \cite{Kim2021} }} & \multicolumn{1}{c}{\textcolor[rgb]{ 0,  0,  0}{\textbf{MIC \cite{MIC}}}} & \multicolumn{1}{c}{\textcolor[rgb]{ 0,  0,  0}{\textbf{CIA\_UDA \cite{CIA_UDA}}}} & \multicolumn{1}{c}{\textcolor[rgb]{ 0,  0,  0}{\textbf{UDA\_ME\_BS \cite{ME_BS}}}} & \multicolumn{1}{c}{\textbf{Ours}} \\
    \midrule
    Surface water & 0.1474 & 0.4721 & 0.4946 & 0.347 & \textcolor[rgb]{ 0,  0,  0}{0.4789} & \textcolor[rgb]{ 0,  0,  0}{0.4909} & \textcolor[rgb]{ 0,  0,  0}{0.1477} & 0.5138 \\
    Street & 0.1110 & 0.2172 & 0.3176 & 0.2151 & \textcolor[rgb]{ 0,  0,  0}{0.3179} & \textcolor[rgb]{ 0,  0,  0}{0.2390} & \textcolor[rgb]{ 0,  0,  0}{0.0546} & 0.3207 \\
    Urban Fabric & 0.3705 & 0.406 & 0.6428 & 0.4573 & \textcolor[rgb]{ 0,  0,  0}{0.6392} & \textcolor[rgb]{ 0,  0,  0}{0.4527} & \textcolor[rgb]{ 0,  0,  0}{0.3879} & 0.6476 \\
    Industrial, commercial and transport & 0.3841 & 0.4301 & 0.7317 & 0.5373 & \textcolor[rgb]{ 0,  0,  0}{0.7072} & \textcolor[rgb]{ 0,  0,  0}{0.5274} & \textcolor[rgb]{ 0,  0,  0}{0.3919} & 0.7376 \\
    Mine, dump, and construction sites & 0.0060 & 0.4214 & 0.5696 & 0.3872 & \textcolor[rgb]{ 0,  0,  0}{0.5583} & \textcolor[rgb]{ 0,  0,  0}{0.5006} & \textcolor[rgb]{ 0,  0,  0}{0.1134} & 0.5949 \\
    Artificial, vegetated areas & 0.1977 & 0.3215 & 0.6568 & 0.4019 & \textcolor[rgb]{ 0,  0,  0}{0.6463} & \textcolor[rgb]{ 0,  0,  0}{0.4243} & \textcolor[rgb]{ 0,  0,  0}{0.2307} & 0.6615 \\
    Arable Land & 0.0151 & 0.6091 & 0.8372 & 0.5975 & \textcolor[rgb]{ 0,  0,  0}{0.8385} & \textcolor[rgb]{ 0,  0,  0}{0.5151} & \textcolor[rgb]{ 0,  0,  0}{0.2438} & 0.8382 \\
    Permanent Crops & 0.0000 & 0.0636 & 0.2136 & 0.1064 & \textcolor[rgb]{ 0,  0,  0}{0.2168} & \textcolor[rgb]{ 0,  0,  0}{0.2447} & \textcolor[rgb]{ 0,  0,  0}{0.0151} & 0.2358 \\
    Pastures & 0.0068 & 0.5996 & 0.626 & 0.5108 & \textcolor[rgb]{ 0,  0,  0}{0.6191} & \textcolor[rgb]{ 0,  0,  0}{0.5005} & \textcolor[rgb]{ 0,  0,  0}{0.1295} & 0.6451 \\
    Forests & 0.1634 & 0.5328 & 0.6294 & 0.4778 & \textcolor[rgb]{ 0,  0,  0}{0.6267} & \textcolor[rgb]{ 0,  0,  0}{0.5282} & \textcolor[rgb]{ 0,  0,  0}{0.2832} & 0.6247 \\
    Shrub & 0.0196 & 0.4053 & 0.5349 & 0.4402 & \textcolor[rgb]{ 0,  0,  0}{0.5180} & \textcolor[rgb]{ 0,  0,  0}{0.3812} & \textcolor[rgb]{ 0,  0,  0}{0.1090} & 0.5404 \\
    Open spaces with no vegetation & 0.0000 & 0.0319 & 0     & 0.0456 & \textcolor[rgb]{ 0,  0,  0}{0.0127} & \textcolor[rgb]{ 0,  0,  0}{0.1128} & \textcolor[rgb]{ 0,  0,  0}{0.0029} & 0.0217 \\
    Inland wetlands & 0.0040 & 0.2691 & 0.4198 & 0.2551 & \textcolor[rgb]{ 0,  0,  0}{0.4794} & \textcolor[rgb]{ 0,  0,  0}{0.3426} & \textcolor[rgb]{ 0,  0,  0}{0.0723} & 0.4503 \\
    \midrule
    BOA(Avg) & 0.1530 & 0.5232 & \textbf{0.64} & 0.4762 & \textcolor[rgb]{ 0,  0,  0}{0.6254} & \textcolor[rgb]{ 0,  0,  0}{0.4997} & \textcolor[rgb]{ 0,  0,  0}{0.2256} & 0.6381 \\
    BOA(Std) & 0.0158 & 0.0134 & 0.0197 & 0.0689 & \textcolor[rgb]{ 0,  0,  0}{0.0237} & \textcolor[rgb]{ 0,  0,  0}{0.0674} & \textcolor[rgb]{ 0,  0,  0}{0.0436} & 0.0208 \\
    mIoU (Avg) & 0.0646 & 0.2429 & 0.3731 & 0.2393 & \textcolor[rgb]{ 0,  0,  0}{0.3706} & \textcolor[rgb]{ 0,  0,  0}{0.2655} & \textcolor[rgb]{ 0,  0,  0}{0.1006} & \textbf{0.3835} \\
    mIoU (Std) & 0.0088 & 0.024 & 0.0148 & 0.0336 & \textcolor[rgb]{ 0,  0,  0}{0.0179} & \textcolor[rgb]{ 0,  0,  0}{0.0371} & \textcolor[rgb]{ 0,  0,  0}{0.0341} & 0.0161 \\
    mF1 (Avg) & 0.1097 & 0.3677 & 0.5134 & 0.3676 & \textcolor[rgb]{ 0,  0,  0}{0.5122} & \textcolor[rgb]{ 0,  0,  0}{0.4046} & \textcolor[rgb]{ 0,  0,  0}{0.1679} & \textbf{0.5255} \\
    mF1 (Std) & 0.0112 & 0.0288 & 0.0174 & 0.0448 & \textcolor[rgb]{ 0,  0,  0}{0.0203} & \textcolor[rgb]{ 0,  0,  0}{0.0496} & \textcolor[rgb]{ 0,  0,  0}{0.0519} & 0.0186 \\
    \end{tabular}%
  }
  \label{tab:Germany}%
\end{table}%


\cref{tab:Germany} presents a performance comparison of our approach on the C2Seg-AB dataset against a zero-shot baseline, where the Prithvi model was fine-tuned exclusively on the source domain. Additionally, we compare our method with {six}  established  domain adaptation methods. GDA \cite{CVPR2024} tackles DA in two steps: (1) MAE-based pre-training of a RS foundation model on the target domain using SLR adapters, followed by (2) supervised fine-tuning with a limited set of labeled samples from the target domain. PCS \cite{PCS} introduces a single-stage framework combining representation learning and domain alignment using a few labeled source samples, primarily for image classification. To adapt PCS for dense, pixel-wise classification, we computed in-domain self-supervision loss, $\mathcal{L}_{InSelf}$, and cross-domain instance-prototype loss, $\mathcal{L}_{CrossSelf}$, at the pixel level on downsampled features maps from both domains, given the high computational demands of the similarity matrix involved in these losses. Similarly, \cite{Kim2021} proposed a cross-domain self-supervised learning approach (CDS) for adaptation with limited source labels. As this method was also designed for image classification, we calculated in-domain instance discrimination loss, $\mathcal{L}_{I-ID}$, and cross-domain matching loss, $\mathcal{L}_{CDM}$, at the pixel level on downsampled features maps from both domains. { The MIC method \cite{MIC} introduces a Masked Image Consistency (MIC) mechanism to improve unsupervised domain adaptation (UDA) by leveraging spatial context relationships within the target domain as auxiliary cues for robust visual recognition. Specifically, MIC enforces consistency between the predictions of masked target images - where random patches are occluded - and the corresponding pseudo-labels generated from the complete image using an exponential moving average (EMA) teacher. Similarly, UDA\_ME\_BS
 \cite{ME_BS} introduces a self-supervised teacher-student framework incorporating two enhancement techniques: mask-enhanced class mix (MECM) and scale-based rare class sampling (SRCS). MECM applies a high proportion of masks to mixed images constructed from both source and target domain samples, encouraging the model to infer the semantic content of masked regions from surrounding context. Meanwhile, SRCS increases the sampling frequency of small-scale rare classes, addressing class imbalance and improving model robustness. Finally, CIA\_UDA
\cite{CIA_UDA} introduces an integer programming mechanism to model category-level relationships between the source and target domains. This approach employs cosine similarity-based category alignment to enhance the adaptation process by facilitating inter-domain category correspondences.}

{
For evaluation, we utilized the official source code provided by each method, except for UDA\_ME\_BS, for which we employed our own implementation, as its official code has not been released yet.} To ensure a balanced comparison, we used the same model architecture for all methods, as described in the ``Network Architecture" subsection, except for GDA, where we used the recommended SLR adapters. Additionally, for uniformity, an equal number of labeled target samples were used for fine-tuning across methods excluding the zero-shot method, even when fine-tuning was not originally part of the method, as with { PCS, GDS, MIC, CIA\_UDA, and UDA\_ME\_BS.} Specifically, 2,000 samples (pixels) per class were selected for fine-tuning in this experiment.

The experimental results show that our approach significantly outperforms the baseline across three widely recognized evaluation metrics for semantic segmentation \cite{GU_27}: F1-score, Mean Accuracy (MA), and mean Intersection over Union (mIoU), demonstrating its effectiveness in addressing the DA challenge. Our method outperforms GDA across all metrics by a notable margin. This improvement can likely be attributed to our framework's use of ViT’s self-attention mechanism, which learns more generalizable features by employing a concatenated source-target sequence for target domain reconstruction, unlike GDA, which uses only unmasked target patches. Furthermore, our approach better handles unlabeled pixels from the target domain through soft pseudo-label fine-tuning. While PCS demonstrates slightly better performance on the MA metric, our method achieves superior results on both mF1 and mIoU metrics. Our method also consistently outperforms CDS across all metrics, likely because these methods, relying heavily on clustering and similarity matching, are more vulnerable to domain shift issues. Consequently, samples from different classes across domains may be mistakenly clustered together, while samples of the same class from different domains may be inaccurately mapped to distant clusters.
{ Our method also demonstrates superior performance compared to MIC, as its masking mechanism does not effectively facilitate learning transferable features from the source domain for the segmentation task in the target domain, unlike our auxiliary MAE-based approach. The class mix strategy, which forms the core of UDA\_ME\_BS, appears to struggle with remote sensing data due to its inherent challenges, including variations in scale, the presence of multiple classes, and significant spatial and spectral diversity. Finally, our method outperforms CIA\_UDA, as its cosine similarity-based category alignment-the fundamental component of CL-can lead to negative transfer, particularly when the model is overconfident in incorrect predictions. Moreover, the reliance on downsampling for computing the assignment matrix may result in the loss of valuable information, further limiting its effectiveness in domain adaptation.} \\

\noindent \textbf{FLAIR Results}
\begin{table}[htbp]
  \centering
 \caption{FLAIR Results-upper section shows the F1 scores by class, and the lower section details the mean and standard deviation for MA, mIoU, and mF1 metrics.}
  \resizebox{\columnwidth}{!}
  {
    \begin{tabular}{lrrrrrrrr}
    \toprule
    \multicolumn{1}{c}{\textbf{Classes}} & \multicolumn{1}{c}{\textbf{Zero Shot}} & \multicolumn{1}{c}{\textbf{GDA \cite{CVPR2024}}} & \multicolumn{1}{c}{\textbf{PCS \cite{PCS}}} & \multicolumn{1}{c}{\textbf{CDS \cite{Kim2021} }} & \multicolumn{1}{c}{\textcolor[rgb]{ 0,  0,  0}{\textbf{MIC \cite{MIC}}}} & \multicolumn{1}{c}{\textcolor[rgb]{ 0,  0,  0}{\textbf{CIA\_UDA \cite{CIA_UDA}}}} & \multicolumn{1}{c}{\textcolor[rgb]{ 0,  0,  0}{\textbf{UDA\_ME\_BS \cite{ME_BS}}}} & \multicolumn{1}{c}{\textbf{Ours}} \\
    \midrule
    building & 0.5203 & 0.3390 & 0.8265 & 0.5333 & \textcolor[rgb]{ 0,  0,  0}{0.8322} & \textcolor[rgb]{ 0,  0,  0}{0.6391} & \textcolor[rgb]{ 0,  0,  0}{0.5306} & 0.8506 \\
    pervious surface & 0.0554 & 0.5882 & 0.5750 & 0.2950 & \textcolor[rgb]{ 0,  0,  0}{0.6616} & \textcolor[rgb]{ 0,  0,  0}{0.4342} & \textcolor[rgb]{ 0,  0,  0}{0.1551} & 0.6985 \\
    impervious surface & 0.5815 & 0.4040 & 0.8369 & 0.5576 & \textcolor[rgb]{ 0,  0,  0}{0.8472} & \textcolor[rgb]{ 0,  0,  0}{0.6607} & \textcolor[rgb]{ 0,  0,  0}{0.6260} & 0.8665 \\
    water & 0     & 0.0307 & 0.0257 & 0.0112 & \textcolor[rgb]{ 0,  0,  0}{0.0270} & \textcolor[rgb]{ 0,  0,  0}{0.0794} & \textcolor[rgb]{ 0,  0,  0}{0.0159} & 0.0421 \\
    deciduous & 0.3164 & 0.5754 & 0.6725 & 0.4542 & \textcolor[rgb]{ 0,  0,  0}{0.7271} & \textcolor[rgb]{ 0,  0,  0}{0.5746} & \textcolor[rgb]{ 0,  0,  0}{0.3380} & 0.7528 \\
    brushwood & 0.0591 & 0.2179 & 0.5537 & 0.3319 & \textcolor[rgb]{ 0,  0,  0}{0.5958} & \textcolor[rgb]{ 0,  0,  0}{0.4240} & \textcolor[rgb]{ 0,  0,  0}{0.1089} & 0.6226 \\
    herbaceous vegetation & 0.6717 & 0.2591 & 0.7435 & 0.4618 & \textcolor[rgb]{ 0,  0,  0}{0.7641} & \textcolor[rgb]{ 0,  0,  0}{0.4500} & \textcolor[rgb]{ 0,  0,  0}{0.5522} & 0.7945 \\
    agricultural land & 0     & 0.6271 & 0.9111 & 0.7262 & \textcolor[rgb]{ 0,  0,  0}{0.9433} & \textcolor[rgb]{ 0,  0,  0}{0.8487} & \textcolor[rgb]{ 0,  0,  0}{0.2489} & 0.9776 \\
    plowed land & 0     & 0.8135 & 0.8625 & 0.8418 & \textcolor[rgb]{ 0,  0,  0}{0.8833} & \textcolor[rgb]{ 0,  0,  0}{0.8710} & \textcolor[rgb]{ 0,  0,  0}{0.5655} & 0.9268 \\
    swimming pool & 0     & 0.0927 & 0.0424 & 0.0540 & \textcolor[rgb]{ 0,  0,  0}{0.0659} & \textcolor[rgb]{ 0,  0,  0}{0.1681} & \textcolor[rgb]{ 0,  0,  0}{0.0686} & 0.0950 \\
    greenhouse & 0.0014 & 0.0913 & 0.0000 & 0.0432 & \textcolor[rgb]{ 0,  0,  0}{0.1924} & \textcolor[rgb]{ 0,  0,  0}{0.1759} & \textcolor[rgb]{ 0,  0,  0}{0.0209} & 0.2429 \\
    \midrule
    BOA(Avg) & 0.228 & 0.6112 & 0.7262 & 0.6858 & \textcolor[rgb]{ 0,  0,  0}{0.8162} & \textcolor[rgb]{ 0,  0,  0}{0.6936} & \textcolor[rgb]{ 0,  0,  0}{0.3962} & \textbf{0.8844} \\
    BOA(Std) & 0.0252 & 0.0318 & 0.0087 & 0.0264 & \textcolor[rgb]{ 0,  0,  0}{0.0692} & \textcolor[rgb]{ 0,  0,  0}{0.0811} & \textcolor[rgb]{ 0,  0,  0}{0.0720} & 0.0428 \\
    mIoU (Avg) & 0.1401 & 0.2694 & 0.4510 & 0.2808 & \textcolor[rgb]{ 0,  0,  0}{0.4938} & \textcolor[rgb]{ 0,  0,  0}{0.3666} & \textcolor[rgb]{ 0,  0,  0}{0.2026} & \textbf{0.5280} \\
    mIoU (Std) & 0.0202 & 0.0434 & 0.0180 & 0.0156 & \textcolor[rgb]{ 0,  0,  0}{0.0350} & \textcolor[rgb]{ 0,  0,  0}{0.0410} & \textcolor[rgb]{ 0,  0,  0}{0.0499} & 0.0223 \\
    mF1 (Avg) & 0.2005 & 0.3672 & 0.5500 & 0.3918 & \textcolor[rgb]{ 0,  0,  0}{0.5945} & \textcolor[rgb]{ 0,  0,  0}{0.4842} & \textcolor[rgb]{ 0,  0,  0}{0.2937} & \textbf{0.6245} \\
    mF1 (Std) & 0.0243 & 0.0443 & 0.0133 & 0.0171 & \textcolor[rgb]{ 0,  0,  0}{0.0349} & \textcolor[rgb]{ 0,  0,  0}{0.0435} & \textcolor[rgb]{ 0,  0,  0}{0.0675} & 0.0264 \\
    \end{tabular}%
  }
   \label{tab:Flair}%
\end{table}%

\cref{tab:Flair} presents the experimental results on the FLAIR dataset, using 600 samples per class for DA. The results highlight the superior performance of our proposed method across all metrics. In this experiment, we examined a scenario where the plowed land and water classes are absent from the source domain, creating an open-set DA challenge in which the source domain classes are a subset of those in the target domain. Focusing on these two classes, our method demonstrates the best performance compared to other approaches, underscoring its effectiveness in open-set scenarios. 
A qualitative visual comparison using inference (segmentation) maps for both datasets is provided in the supplementary material.

\subsection{Ablation Study and Analysis}
\vspace{-6pt}
\begin{table}[htbp]
  \centering
  \caption{Contribution of each component in our proposed framework to overall performance on the FLAIR dataset.}
  \resizebox{.9\linewidth}{!}{%
    \begin{tabular}{lrrrr}
      \toprule
      \multicolumn{1}{c}{\textbf{Classes}} & \multicolumn{1}{c}{$\mathcal{L}_{Seg}$} & \multicolumn{1}{c}{$\mathcal{L}_{Seg}+\mathcal{L}_{DA}$} & \multicolumn{1}{c}{$\mathcal{L}_{Seg}+\mathcal{L}_{MAE}$} & \multicolumn{1}{c}{$\mathcal{L}_{Seg}+\mathcal{L}_{DA}+\mathcal{L}_{MAE}$ (Ours)} \\
      \midrule
      building & 0.5844 & 0.8276 & 0.8464 & 0.8506 \\
      pervious surface & 0.3743 & 0.6494 & 0.6593 & 0.6985 \\
      impervious surface & 0.6770 & 0.8465 & 0.8576 & 0.8665 \\
      water & 0.0938 & 0.0983 & 0.0633 & 0.0421 \\
      deciduous & 0.5370 & 0.7006 & 0.7392 & 0.7528 \\
      brushwood & 0.3677 & 0.6133 & 0.6133 & 0.6226 \\
      herbaceous vegetation & 0.3336 & 0.7664 & 0.7906 & 0.7945 \\
      agricultural land & 0.6968 & 0.9522 & 0.9640 & 0.9776 \\
      plowed land & 0.8586 & 0.8793 & 0.8914 & 0.9268 \\
      swimming pool & 0.2048 & 0.0673 & 0.0983 & 0.0950 \\
      greenhouse & 0.0649 & 0.0577 & 0.2699 & 0.2429 \\
      \midrule
      MA(Avg) & 0.6109 & 0.7921 & 0.8631 & 0.8844 \\
      MA(Std) & 0.0577 & 0.0519 & 0.0688 & 0.0428 \\
      mIoU (Avg) & 0.3239 & 0.4878 & 0.5156 & 0.5280 \\
      mIoU (Std) & 0.0347 & 0.0226 & 0.0283 & 0.0223 \\
      mF1 (Avg) & 0.4357 & 0.5871 & 0.6176 & 0.6245 \\
      mF1 (Std) & 0.0372 & 0.0230 & 0.0301 & 0.0264 \\
    \end{tabular}%
  }
  \label{tab:Abilation_Flair}%
\end{table}%

\cref{tab:Abilation_Flair} demonstrates the effectiveness of each component in our proposed method on the FLAIR dataset. {Unlike the zero-shot setting presented in the first column of the table, which involves fine-tuning exclusively on the source domain, the second column reports the baseline performance, where standard fine-tuning is applied using a limited number of labeled target domain samples in conjunction with the source domain.} Adding each component to this baseline progressively enhances performance, with no degradation observed. Notably, the combination of $\mathcal{L}_{DA}$ and $\mathcal{L}_{MAE}$ exhibits a synergistic effect: $\mathcal{L}_{MAE}$ functions as a weighting mechanism, adjusting in alignment with the segmentation model’s confidence in the unlabeled target domain pixels, as detailed in the supplementary material. Meanwhile, $\mathcal{L}_{DA}$ directly encourages higher confidence in these unlabeled pixels.

\subsection{MAE-Based Generative Performance}
To assess the generative capability of our model, we conduct an evaluation using MAE-based learning across different data modalities. Specifically, we analyze its ability to reconstruct target domain images given a source-target sequence, highlighting its potential to learn domain-invariant representations. This evaluation is performed at varying target domain masking ratios while keeping source images unmasked. The results demonstrate strong reconstruction performance both spatially and spectrally, underscoring the effectiveness of our approach. Further details and visual assessments can be found in the supplementary material.

\vspace{-6pt}

\section{Conclusion}
This study presents a novel approach that integrates soft-alignment pseudo-labeling with source-to-target MAE generative pre-training to enhance DA in RS. By introducing a pre-training framework adaptable to various data modalities, our method extends the feature extraction capabilities of foundational models in RS. Validation on two datasets with distinct modalities- reflecting variability between source and target domain images spatially, temporally, and in the number of classes, etc-demonstrates the method’s effectiveness in boosting the model's generalization and adaptability across domains. Alongside a rigorous mathematical analysis of MAE-based generative learning’s impact on DA, we empirically validate its strong generative performance for passive optical geospatial imagery acquired from different sensors.
\section*{Acknowledgments}
 This work was supported by the NASA grant $\# 80NSSC22K1163$: Knowledge Transfer for Robust GeoAI Across Space, Sensors and Time via Active Deep Learning. Additionally, this work was completed in part with resources provided by the Research Computing Data Core at the University of Houston.
{
    \small
    \bibliographystyle{ieeenat_fullname}
    \bibliography{main}
}

\clearpage
\setcounter{page}{1}
\setcounter{section}{0}
\renewcommand{\thesection}{\Alph{section}}
\maketitlesupplementary
\def\x{{\mathbf x}}
\def\L{{\cal L}}
\def\E{{\mathbb{E}}}

\section{Mathematical Insight}
\label{sec:rationale}

\subsection{Notations and Definitions}

\noindent  \textbf{Datasets Notations and Definitions:} Let $Y_i^\tau$ represent the class (as a random variable) at the $i^{th}$ pixel of an image $\mathbf{X}_\tau$ taken from domain $\tau$, where $\tau \in \{S, T\}$ denotes the source or target domain, respectively. The random variable $X_i^\tau$ represents the channel vector at the $i^{th}$ pixel in the image $\mathbf{X}_\tau$, i.e., $X_i^\tau \in \mathbb{R}^{C_{\tau}}$, where $C_{\tau}$ is the number of channels. An observation of of the random variable $Y_i^\tau$ is denoted as $y_i^\tau$, and similarly, an observation of the random variable $X_i^\tau$ is denoted as $x_i^\tau$.

Let the set $\mathbf{X}_S = \{x_i^S \}_{i=1}^{N_S = H_S \times W_S}$ represent the source domain image, where $\mathbf{X}_S \in \mathbb{R}^{C_S \times H_S \times W_S}$ is the set of channel vectors for all pixels in the source image, with $H_S$ and $W_S$ being the height and width of the image, respectively. Similarly, let the set $\mathbf{X}_T = \{x_i^T\}_{i=1}^{N_T = H_T \times W_T}$ represent the target domain image, where $\mathbf{X}_T \in \mathbb{R}^{C_T \times H_T \times W_T}$ is the set of channel vectors for all pixels in the target image.

The labels for the source domain are represented by the set $\mathbf{Y}_S = \{ y_i^S \}_{i=1}^{N_S}$, where we assume the source domain is fully labeled. For the target domain, the labeled samples are represented by the set $\mathbf{Y}_T = \{ y_i^T \}_{i=1}^l$, where $l$ is the number of labeled pixels and $l \ll N_T$.

\noindent  \textbf{Models Defentions :} Let $f_\theta$: the feature extractor parametrized by $\theta$

$h_{\theta_{\text{seg}}}$: the segmentation head parametrized by $\theta_{\text{seg}}$

$g_{\theta_M}$: MAE head parametrized by $\theta_M$

Then we can define the segmentation model and the generative model we have in our proposed method as follows :
\begin{align}
    \mathbb{P}(y_i \mid \mathbf{X}; \theta, \theta_{\text{seg}}) 
    &= h_{\theta_{\text{seg}}}(f_\theta(\mathbf{X}), y_i) \label{eq:D1}
\end{align}
The probability of class $y$ at the $i^{th}$ pixel given the entire image $\mathbf{X}$. By employing mean squared error (MSE) in the generative component, we inherently assume the following model :
\begin{align}
X_i^T &= g_{\theta_M} \left( f_\theta (X_S) \right) + \epsilon, \nonumber \\
& \text{where } \epsilon \sim \mathcal{N}(0, \Sigma), \nonumber \\
& \text{and } \Sigma \in \mathbb{R}^{C_T \times C_T}. \label{eq:D2}
\end{align}

\subsection{Task Learning as Joint Likelihood   }
The segmentation, generative, and  domain adaptation (DA) tasks in the proposed method are governed by minimizing the following loss function:
\begin{align} 
\underset{\theta,\theta_{seg},\theta_{M}}{\mathrm{argmin}} \, \mathcal{L}_{Tot}(\lambda_{DA}, \lambda_{MAE}, \hdots) &= \mathcal{L}_{Seg} + \lambda_{DA} \mathcal{L}_{DA} \nonumber \\
&\quad + \lambda_{MAE} \mathcal{L}_{MAE} \label{total_loss} 
\end{align}
 Here, $\L_{Seg}$ denotes the multi-class cross-entropy loss, computed on the labeled source domain and a limited number of labeled target domain samples. The $\L_{DA}$ term facilitates domain alignment between the source and target domains, while $\L_{MAE}$ represents the mean squared error (MSE) used in MAE method.

For clarity in this mathematical insight, we focus solely on the impact of MAE learning on the segmentation task. Therefore, we set \(\lambda_{DA}=0\) and \(\lambda_{MAE}=1\) in \Cref{total_loss}. Hence 
\begin{align}
\underset{\theta, \theta_{seg}, \theta_{M}}{\mathrm{argmin}} \, \mathcal{L}_{Tot}(0,1,\hdots) 
&= \underset{\theta, \theta_{seg}, \theta_{M}}{\mathrm{argmax}} \, J(\theta, \theta_{seg}, \theta_{M}) 
\label{eq:equivilant}
\end{align}
Where $J(\theta,\theta_{seg},\theta_{M})$ is the joint probabilty  of the \textit{observed}  source domain labels, $\mathbf{Y}_S$, the few  \textit{observed} labels from the target domain, $\mathbf{Y}_T$, and the \textit{observed} target domain image $\mathbf{X}_T$, conditioned on the \textit{observed} source domain image, $\mathbf{X}_S$, and parametrized by the model parameters :
\begin{align}
    J(\theta, \theta_{\text{seg}}, \theta_{M}) = 
    \mathbb{P}(\mathbf{Y}_S, \mathbf{Y}_T, \mathbf{X}_T \mid \mathbf{X}_S; \theta, \theta_{\text{seg}}, \theta_{M}) 
    \label{eq:Joint_distrution}
\end{align}
The joint distribution in \Cref{eq:Joint_distrution} can be factorized as follows:


\begin{align}
\mathbb{P}(\mathbf{Y}_S, \mathbf{Y}_T, \mathbf{X}_T 
\mid \mathbf{X}_S; \theta, \theta_{\text{seg}}, \theta_{M}) 
&= 
\mathbb{P}(\mathbf{Y}_S \mid \mathbf{Y}_T, \mathbf{X}_T, \mathbf{X}_S) \nonumber \\
&\quad \times 
\mathbb{P}(\mathbf{Y}_T \mid \mathbf{X}_T, \mathbf{X}_S) \nonumber \\
&\quad \times 
\mathbb{P}(\mathbf{X}_T \mid \mathbf{X}_S),
\label{eq:Joint_distribution}
\end{align}

Assuming the conditional independence assumption holds for the output quantities given the inputs, we have \( Y^\tau_i \perp \zeta \mid \mathbf{X_\tau} \) for the segmentation model, where \( \zeta \) represents any variable other than \( Y^\tau_i \). Likewise, for the MAE model, the condition \( X_i^T \perp \zeta \mid \mathbf{X}_S \) applies for all \( \zeta \) not equal to \( X_i^T \). Under these assumptions, the joint distribution in \Cref{eq:Joint_distribution} simplifies as follows:

\begin{align}
\mathbb{P}(\mathbf{Y}_S, \mathbf{Y}_T, \mathbf{X}_T \mid \mathbf{X}_S; \theta, \theta_{\text{seg}}, \theta_{M}) 
&= \mathbb{P}(\mathbf{Y}_S \mid \mathbf{X}_S; \theta, \theta_{\text{seg}}) \nonumber \\
&\quad \times \mathbb{P}(\mathbf{Y}_T \mid \mathbf{X}_T; \theta, \theta_{\text{seg}}) \nonumber \\
&\quad \times \mathbb{P}(\mathbf{X}_T \mid \mathbf{X}_S; \theta, \theta_{M}) \label{eq:Joint_distribution_simplified}
\end{align}

Utilizing the  conditional independence  assumption to write \Cref{eq:Joint_distribution_simplified} on the pixel level :
\begin{align}
J &= \prod_{n \in \{1,\hdots,N_S\}} \mathbb{P}(y_n^s \mid \mathbf{X}_S; \theta, \theta_{\text{seg}}) \nonumber \\
  &\quad \times \prod_{i \in \{1,\hdots,l\}} \mathbb{P}(y_i^T \mid \mathbf{X}_T; \theta, \theta_{\text{seg}}) \nonumber \\
  &\quad \times \prod_{k \in \{1,\hdots,N_T\}} \mathbb{P}(x_k^T \mid \mathbf{X}_S; \theta, \theta_{M}) \label{eq:joint_pixel_level}
\end{align}
\subsubsection{Impact of Unlabeled Target Domain Sample Reconstruction on the Segmentation Task  }
Since our approach incorporates a few labeled samples from the target domain alongside labeled data from the source domain, the impact of these labeled target samples on the segmentation task can be inferred through their direct influence on the decision boundaries of the segmentation model. However, the use of the MAE  loss has enabled the integration of unlabeled samples from the target domain. We are specifically focused on investigating the impact of these unlabeled samples on the segmentation task. To achieve this, we will reformulate \Cref{eq:joint_pixel_level} to distinguish between the labeled and unlabeled pixels from the target domain. Before doing so, it is essential to first highlight the following product that appears in \Cref{eq:joint_pixel_level}.

\begin{align}
\mathbb{P}(y_i^T \mid \mathbf{X}_T; \theta, \theta_{\text{seg}}) 
\mathbb{P}(x_k^T \mid \mathbf{X}_S; \theta, \theta_{M}) \label{eq:important_quantity}
\end{align}

According to the basic rule of conditional probability and the  conditional independence  assumption, Expression (\ref{eq:important_quantity}) can be written as follows:
\begin{align}
    & \mathbb{P}\left( y_i^T, x_k^T \mid \{ x_r^T \}_{r \neq k}, \mathbf{X}_S; \bar{\theta} = [\theta, \theta_{\text{seg}}, \theta_M] \right) \nonumber \\
    & \quad = \mathbb{P}\left( y_i^T \mid \mathbf{X}_T; \theta, \theta_{\text{seg}} \right) \nonumber \\
    & \quad \quad \times \mathbb{P}\left( x_k^T \mid \mathbf{X}_S; \theta, \theta_M \right)
    \label{eq:joint_distribtion_target}
\end{align}

In words,  Expression (\ref{eq:important_quantity}) represents the joint probability of \( Y_i^T \) and \( X_k^T \), parameterized by \( \bar{\theta} = [\theta, \theta_{\text{seg}}, \theta_M] \), conditioned on \( \mathbf{X}_S \) and all channel vectors of \( \mathbf{X}_T \), excluding \( x_k^T \), which we denote as \( \{ x_r^T \}_{r \neq k} \).

Substituting \Cref{eq:joint_distribtion_target} in \Cref{eq:joint_pixel_level} and taking the $log$ we got :
\begin{align}
\log {J} &= \sum_{n \in \{1, \dots, N_S\}} \log \mathbb{P}\left( y_n^S \mid \mathbf{X}_S, \theta, \theta_{\text{seg}} \right) \nonumber \\
&\quad + \sum_{i \in \{1, \dots, l\}} \log \mathbb{P}\left( y_i^T, x_i^T \mid \{ x_r^T \}_{r \neq i}, \mathbf{X}_S, \bar{\theta} \right) \nonumber \\
&\quad + \sum_{k \in \{1, \dots, N_T\} \setminus \{1, \dots, l\}} \log \mathbb{P}\left( x_k^T \mid \mathbf{X}_S, \theta, \theta_M \right).
\label{eq:log_jointlikelhood}
\end{align}

To evaluate how this learning paradigm influences the feature extractor during training, a key component of our proposed method for jointly learning the segmentation and MAE tasks, we will compute the partial derivative with respect to $\theta$. This approach is motivated by gradient-based optimization methods, where the derivative is responsible for updating the model parameters, enabling us to observe how the feature extractor evolves and adapts throughout the training process:
\begin{align}
\frac{\partial J}{\partial \theta} &= 
\sum_{n \in \{1, \dots, N_S\}} \frac{\partial}{\partial \theta} 
\log \mathbb{P}\left( y_n^S \mid \mathbf{X}_S, \theta, \theta_{\text{seg}} \right) \nonumber \\
&+ \sum_{i \in \{1, \dots, l\}} \frac{\partial}{\partial \theta} 
\log \mathbb{P}\left( y_i^T, x_i^T \mid \{ x_r^T \}_{r \neq i}, \mathbf{X}_S, \bar{\theta} \right) \nonumber \\
&+ \sum_{k \in \{1, \dots, N_T\} \setminus \{1, \dots, l\}} \frac{\partial}{\partial \theta} 
\log \mathbb{P}\left( x_k^T \mid \mathbf{X}_S, \theta, \theta_M \right) 
\label{eq:dev-log-joint}
\end{align}

By utilizing the following ``trick " provided by  \cite{Cambridge}: 
\begin{equation}
\begin{aligned}
\forall x, y, \frac{\partial}{\partial \theta} \log \mathbb{P}(x|\theta) 
&= \frac{1}{\mathbb{P}(x|\theta)} \frac{\partial}{\partial \theta} \mathbb{P}(x|\theta) \\
&= \frac{1}{\mathbb{P}(x|\theta)} \frac{\partial}{\partial \theta} \left( \sum_{y} \mathbb{P}(x, y|\theta) \right) \\
&= \frac{1}{\mathbb{P}(x|\theta)} \sum_{y} \frac{\partial}{\partial \theta} \mathbb{P}(x, y|\theta) \\
&= \frac{1}{\mathbb{P}(x|\theta)} \sum_{y} \mathbb{P}(x, y|\theta) 
   \left( \frac{1}{\mathbb{P}(x, y|\theta)} \frac{\partial}{\partial \theta} \mathbb{P}(x, y|\theta) \right) \\
&= \sum_{y} \frac{\mathbb{P}(x, y|\theta)}{\mathbb{P}(x|\theta)} \frac{\partial}{\partial \theta} \log \mathbb{P}(x, y|\theta) \\
&= \sum_{y} \mathbb{P}(y|x, \theta) \frac{\partial}{\partial \theta} \log \mathbb{P}(x, y|\theta).
\end{aligned}
\label{eq:trick}
\end{equation}

Applying the trick in  \Cref{eq:trick} for the last term in \Cref{eq:dev-log-joint}, we got : 
\begin{align}
\frac{\partial{J}}{\partial \theta} = &
\sum_{n \in \{1,\hdots, N_S\}} \frac{\partial}{\partial \theta} 
\log \mathbb{P}\left( y_n^S \mid \mathbf{X}_S, \theta, \theta_{\text{seg}} \right) \nonumber \\
&+ \sum_{i \in \{1,\hdots, l\}} \frac{\partial}{\partial \theta} 
\log \mathbb{P}\left( y_i^T, x_i^T \mid \{ x_r^T \}_{r \neq i}, \mathbf{X}_S, \bar{\theta} \right) \nonumber \\
&+ \sum_{k \in \{1, \dots, N_T\} \setminus \{1, \dots, l\}} 
\sum_{y_k} \mathbb{P}(y_k \mid x_k^T, \mathbf{X}_S; \bar{\theta}) \nonumber \\
& \times \frac{\partial}{\partial \theta} 
\log \mathbb{P}(x_k^T, y_k \mid \mathbf{X}_S; \bar{\theta}) 
\label{eq:joint-trick}
\end{align}

The final term in \Cref{eq:joint-trick} holds particular importance, as it demonstrates how our learning paradigm dynamically adjusts the influence of unlabeled pixels on the classification process for each class. This adjustment is achieved through the expression $\mathbb{P}(y_k \mid x_k^T, \mathbf{X}_S; \bar{\theta})$, which functions as a dynamic weighting mechanism. In the following, we will further explain this dynamic weighting mechanism by relating it to the definitions of our models introduced earlier.

\subsubsection{Dynamic Weighting in terms of the Segmentation Model}
\label{proof:Expectation}
By applying the conditional probability rule and the  conditional independence  assumption, the joint distribution $\mathbb{P}\left( y_k, \{x_r^T \}_{r \neq k} \mid x_k^T; \bar{\theta} \right) $ can be written:

\begin{equation}
\begin{aligned}
\mathbb{P}\left( y_k, \{x_r^T \}_{r \neq k} \mid x_k^T; \bar{\theta} \right) 
&= \mathbb{P}\left( y_k \mid \mathbf{X}_T, \mathbf{X}_S, \bar{\theta} \right) \\
& \times \mathbb{P}\left( \{x_r^T \}_{r \neq k} \mid \mathbf{X}_S, \bar{\theta} \right) \\
&= h(f(\mathbf{X}_T), y_k) \\
&\times \prod_{r \neq k} \mathcal{N}(x_r^T; g(f(\mathbf{X}_S,r)),\Sigma).
\end{aligned}
\label{eq:final_joint}
\end{equation}

\begin{equation}
    \mathbb{P}(y_k \mid x_k^T, \mathbf{X}_S, \bar{\theta})\times \mathcal{N}(x_k^T; g(f(\mathbf{X}_S,k)),\Sigma)
\end{equation}

Now we can direclty link the weight  $\mathbb{P}(y_k \mid x_k^T, \mathbf{X}_S; \bar{\theta})$ to our segmentation model $ h(f(.))$ through marginalising the joint distribution in \Cref{eq:final_joint}

\begin{align}
\mathbb{P}(y_k \mid x_k^T, \mathbf{X}_S, \bar{\theta}) 
&= \int \cdots \int_{\{x_r^T\}_{r \neq k}} 
    h(f(\mathbf{X}_T), y_k) \nonumber \\
& \times \mathbb{P}\left( \{x_r^T \}_{r \neq k} \mid \mathbf{X}_S, \bar{\theta} \right) 
    \,  \prod_{r \neq k} d x_r^T .
\label{eq:weight_segmentation}
\end{align}

In order to write \Cref{eq:weight_segmentation} in a  compact way, we let $\mathbf{Z}=\{x_r^T \}_{r \neq k}$ such that \Cref{eq:weight_segmentation} can be written as follows: 
\begin{align}
\mathbb{P}(y_k \mid x_k^T, \mathbf{X}_S, \bar{\theta}) 
&= \mathbb{E}_{\mathbf{Z}\sim \mathbb{P}(\mathbf{Z} \mid \mathbf{X}_S, \bar{\theta})} 
\left[ h(f(\mathbf{Z}, x_k^T), y_k) \right] \label{eq:final}
\end{align}

\Cref{eq:final} demonstrates that the weighting mechanism is closely linked to the segmentation model's confidence regarding the unlabeled pixels in the target domain. If the model is incorrectly overconfident about these unlabeled pixels, it can adversely affect the learning of the segmentation task. Conversely, when the model is correctly confident about these unlabeled pixels, it enhances the model's generalizability over the target domain by effectively leveraging the unlabeled samples in conjunction with the limited labeled ones.
\section{Implementation Details}
The experiments were conducted using the PyTorch framework with the AdamW optimization algorithm \cite{PRO_REF_45}. A learning rate of $10^{-4}$ was used. Each experiment was repeated ten times for each datasets, with a unique random initialization of the network for each run. In each experiment, validation and testing were performed on datasets sampled from the target domain, and the best model was selected based on validation performance during training. Both hyperparameters, $\lambda_{MAE}$ and $\lambda_{DA}$, were empirically set to a value of 1.

\section{Inference Maps}
\subsection{C2Seg-AB }
\begin{figure*}[htbp]
    \centering
    \setcounter{subfigure}{0}
    
    \subfloat[]{\includegraphics[width=0.3\textwidth]{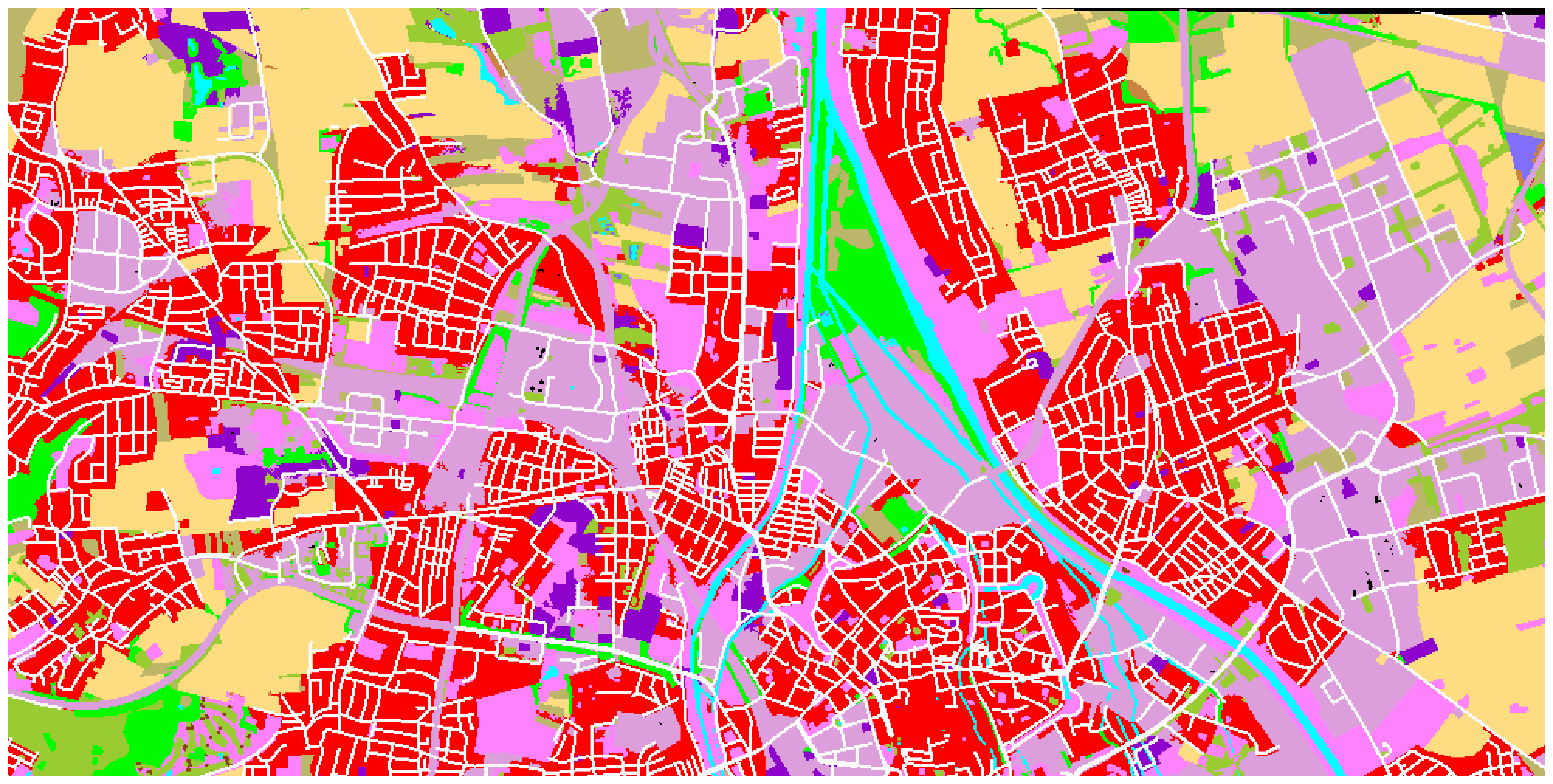}}
    \hfill
    \subfloat[]{\includegraphics[width=0.3\textwidth]{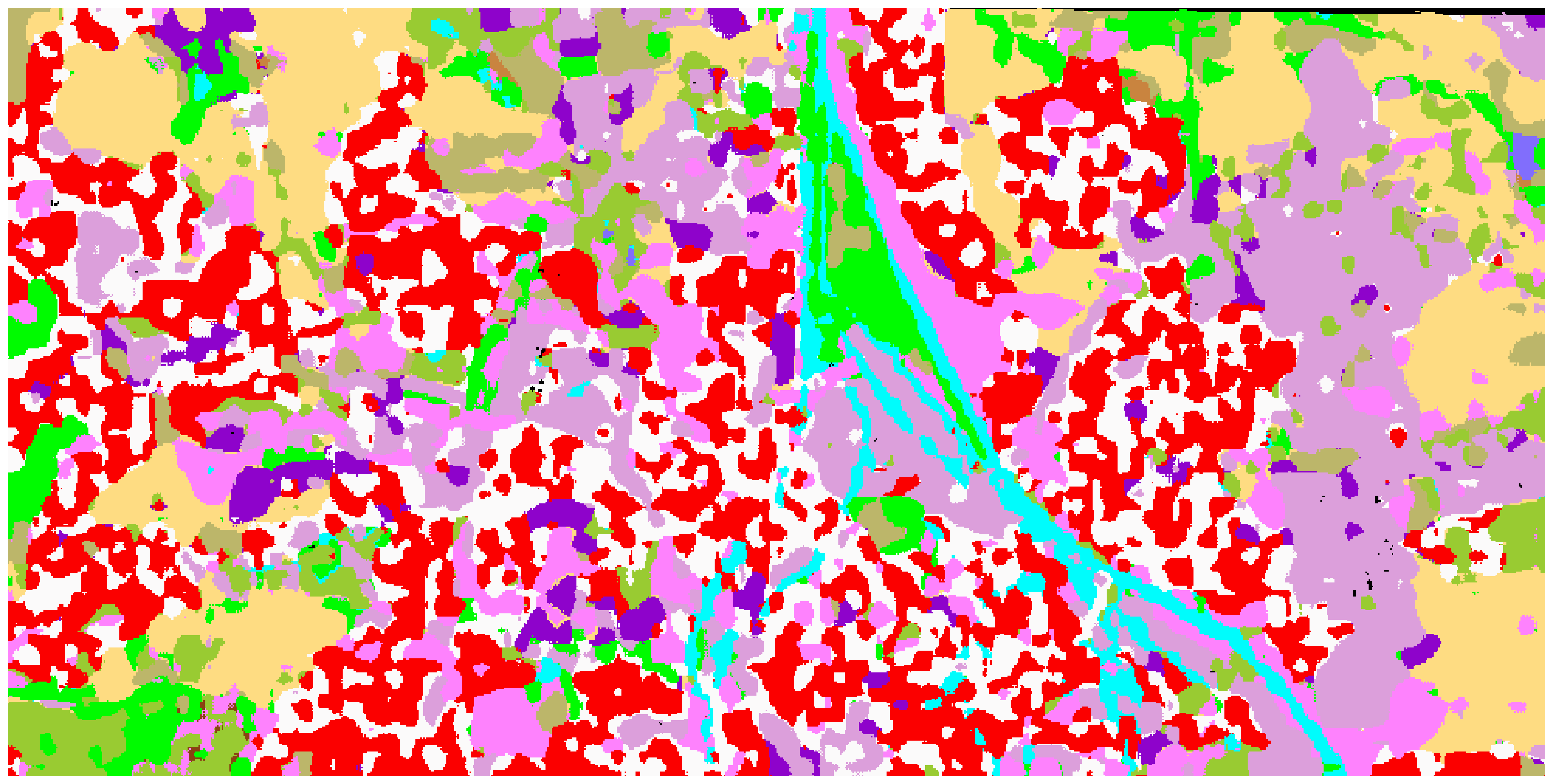}\label{subfig:2}}
    \hfill
    \subfloat[]{\includegraphics[width=0.3\textwidth]{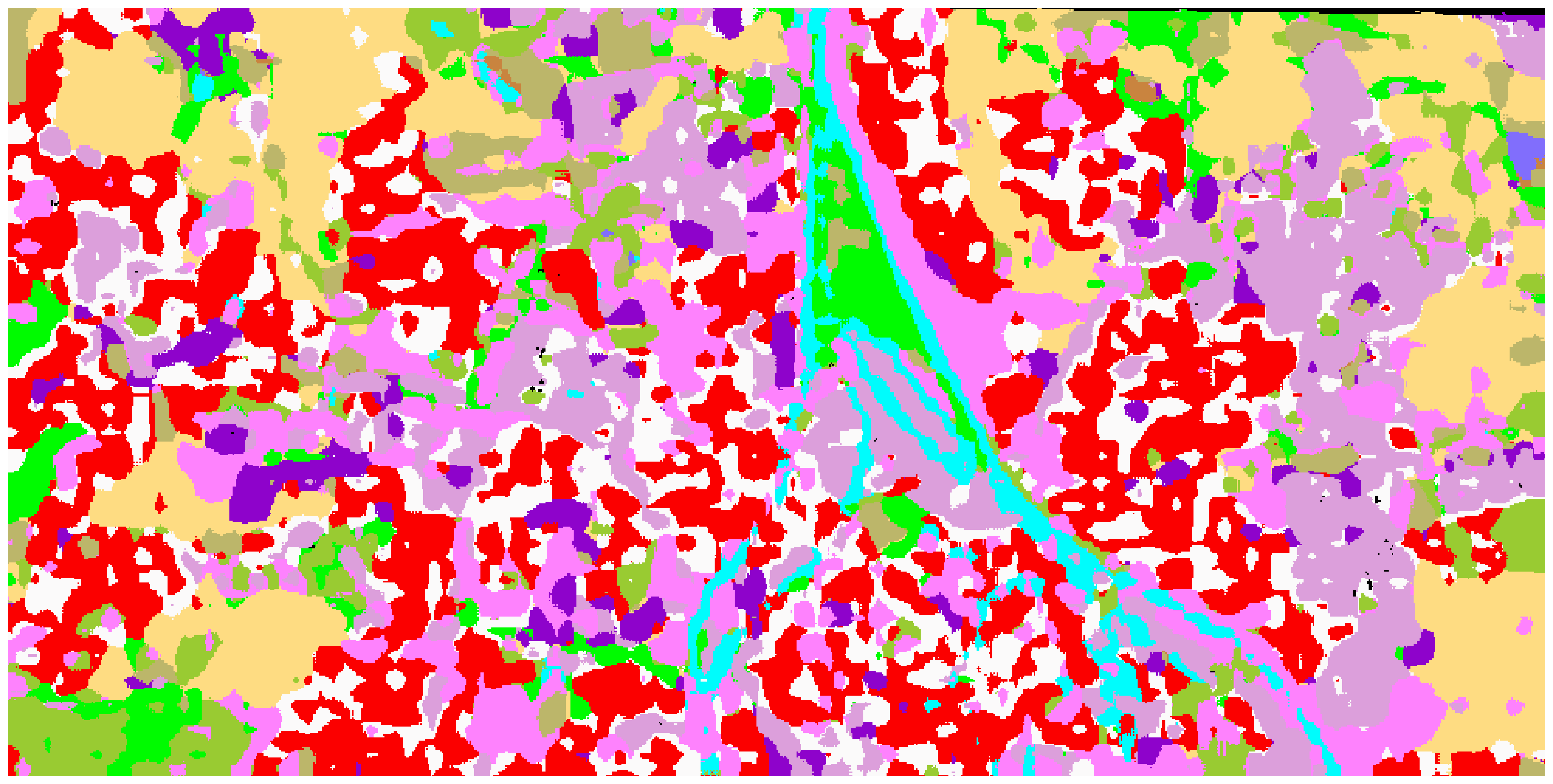}}
    
    \vspace{0.3cm} 

    \subfloat[]{\includegraphics[width=0.3\textwidth]{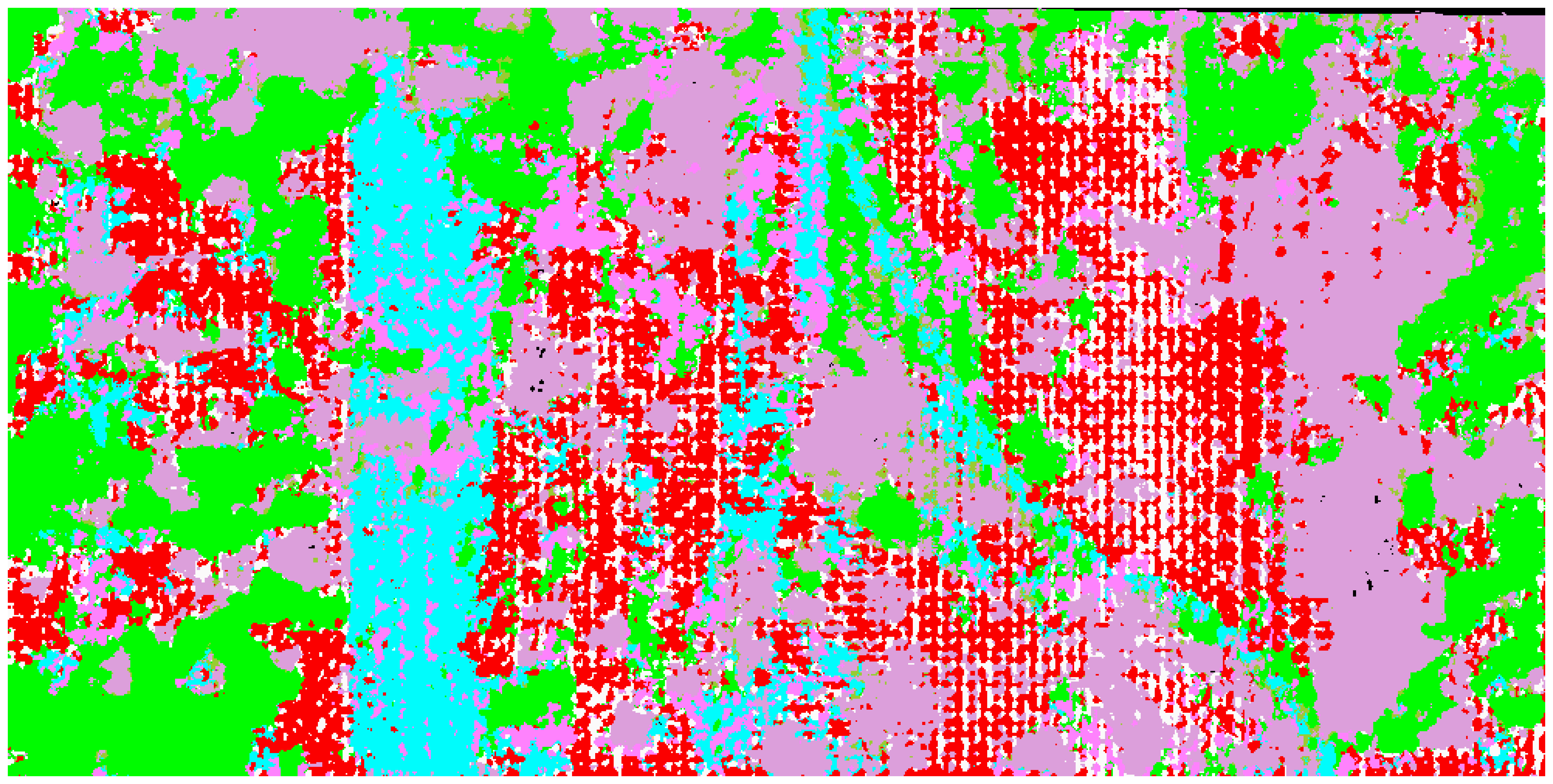}}
    \hfill
    \subfloat[]{\includegraphics[width=0.3\textwidth]{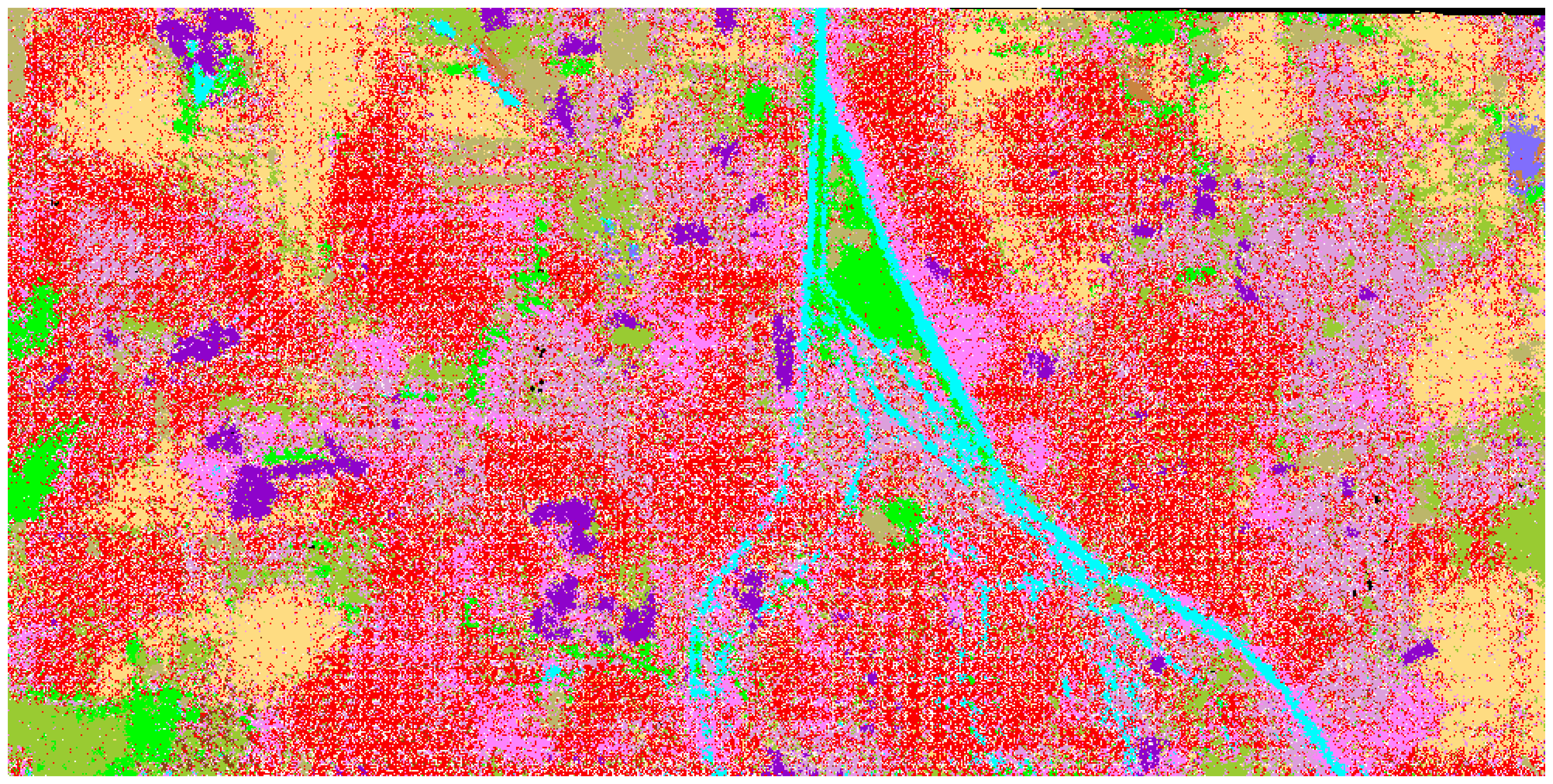}}
    \hfill
    \subfloat[]{\includegraphics[width=0.3\textwidth]{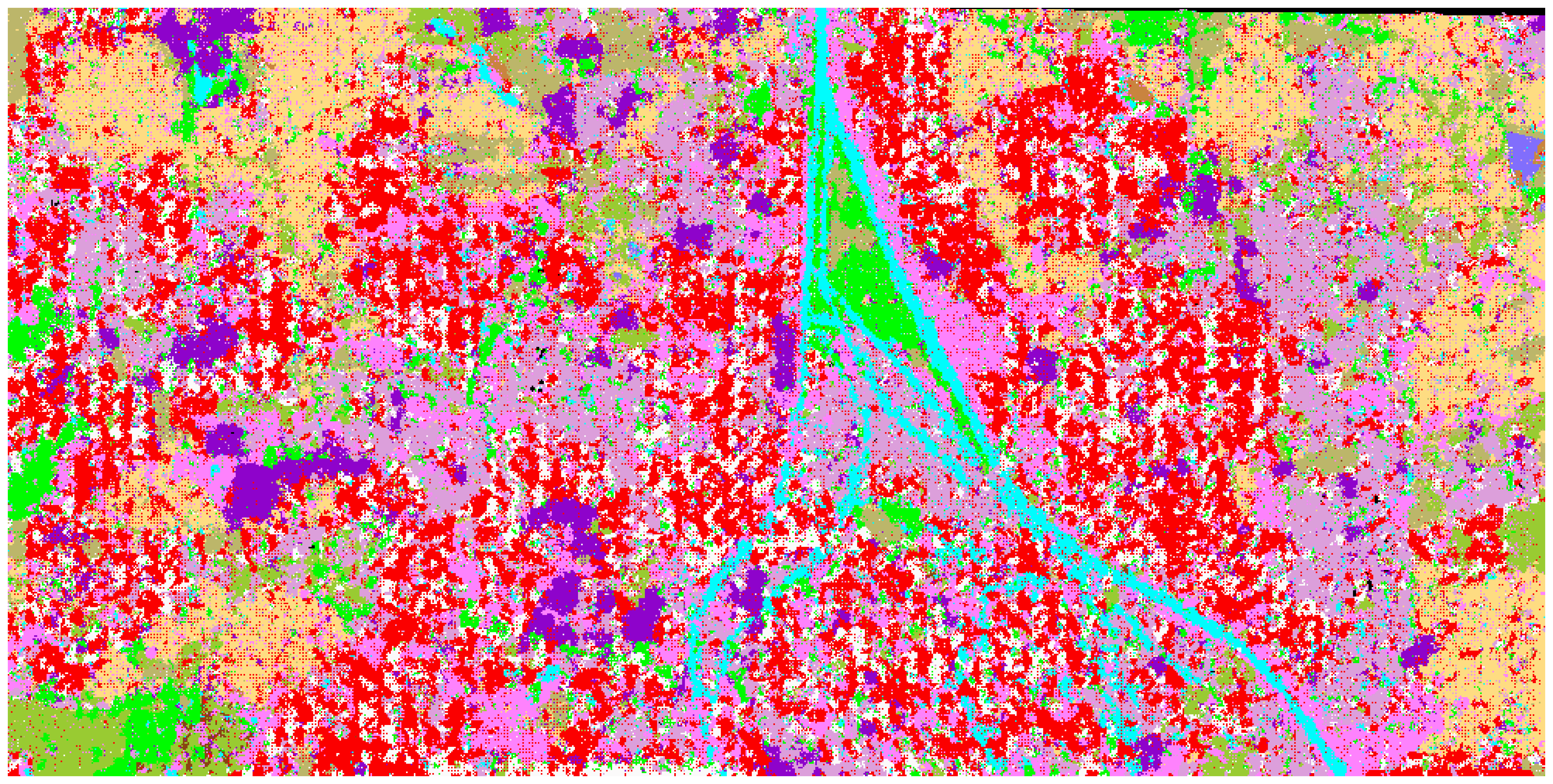}}

    \vspace{0.3cm} 

    \subfloat[]{\includegraphics[width=0.3\textwidth]{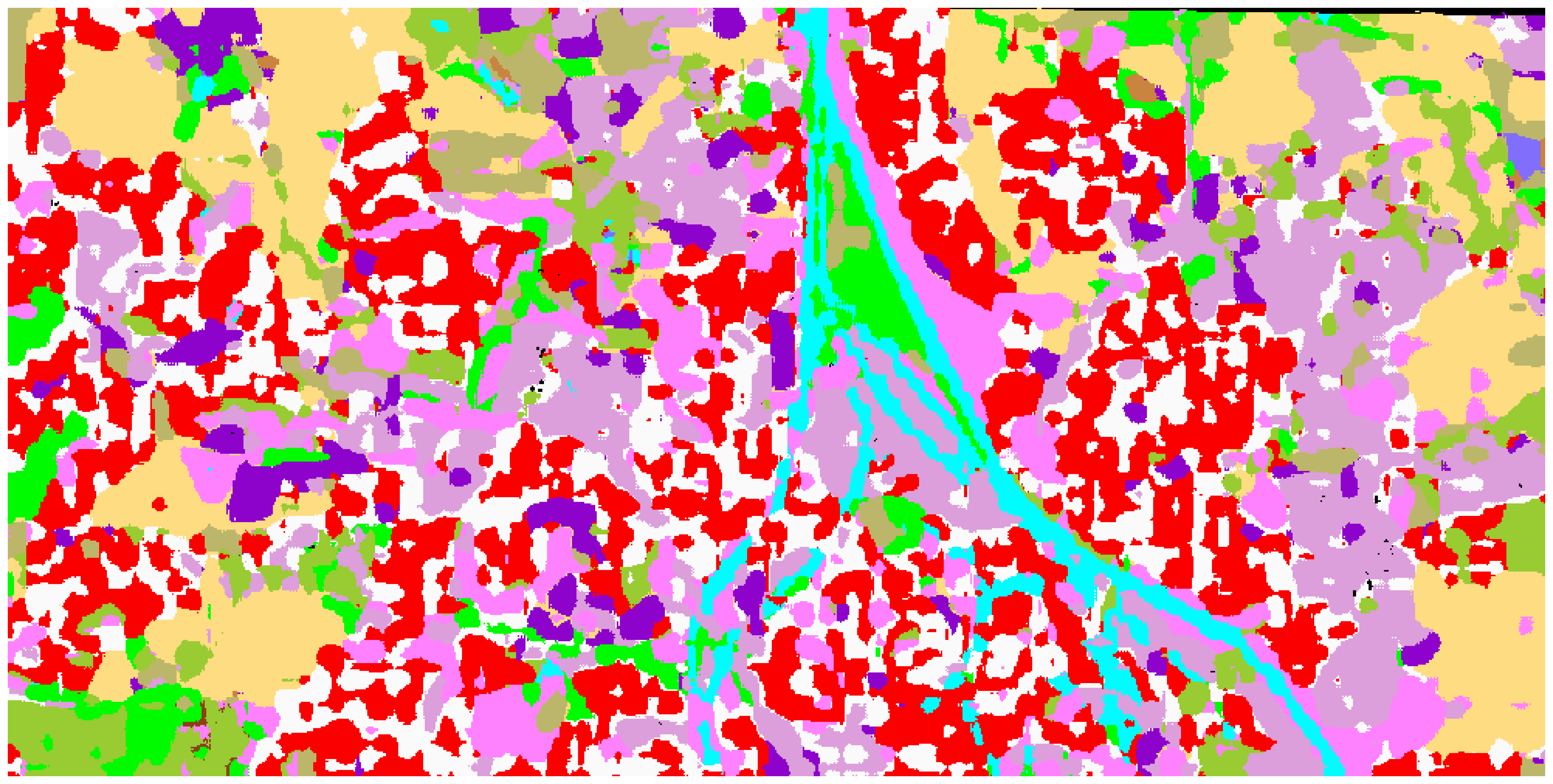}}
    \hfill
    \subfloat[]{\includegraphics[width=0.3\textwidth]{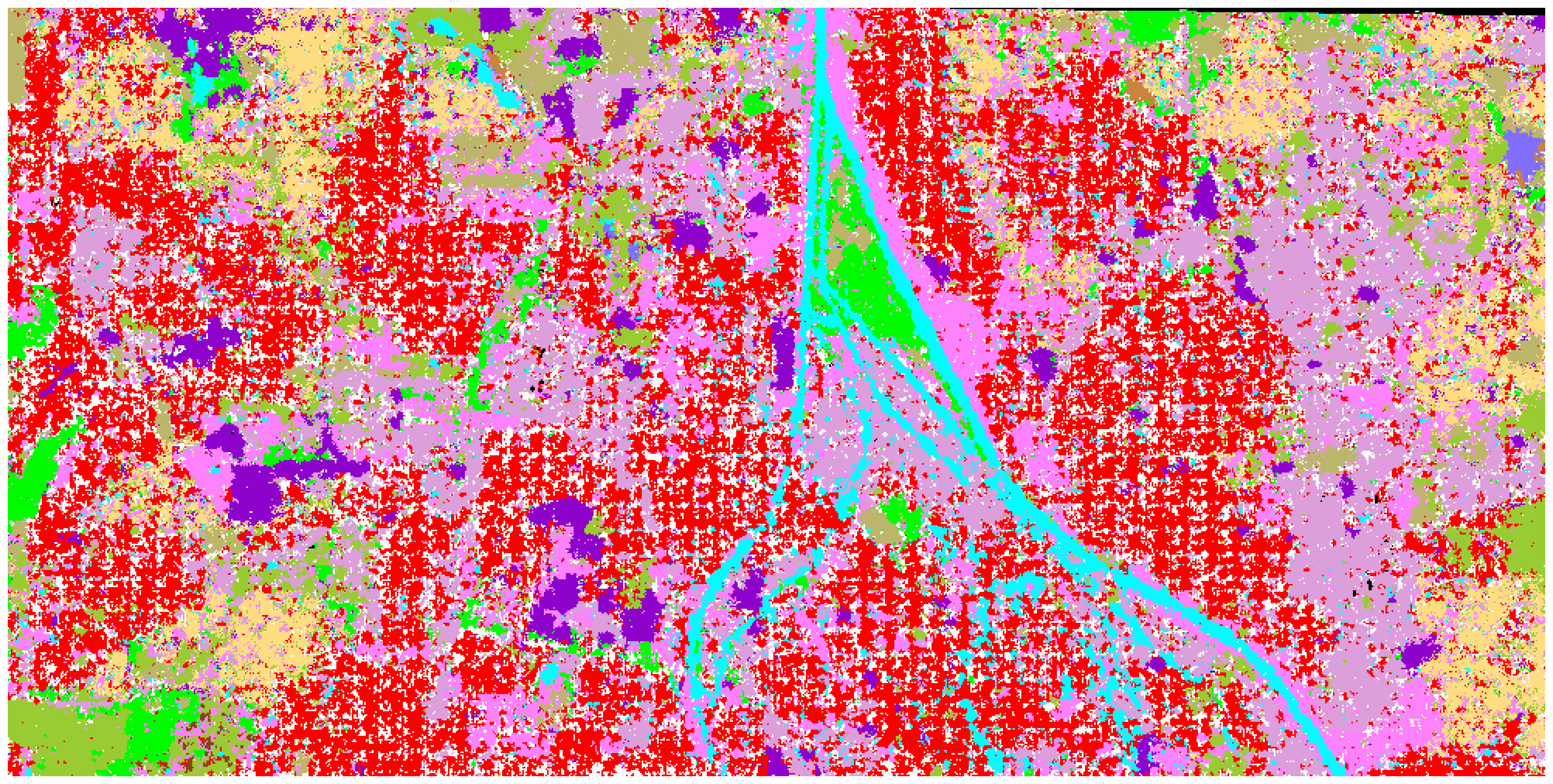}}
    \hfill
    \subfloat[]{\includegraphics[width=0.3\textwidth]{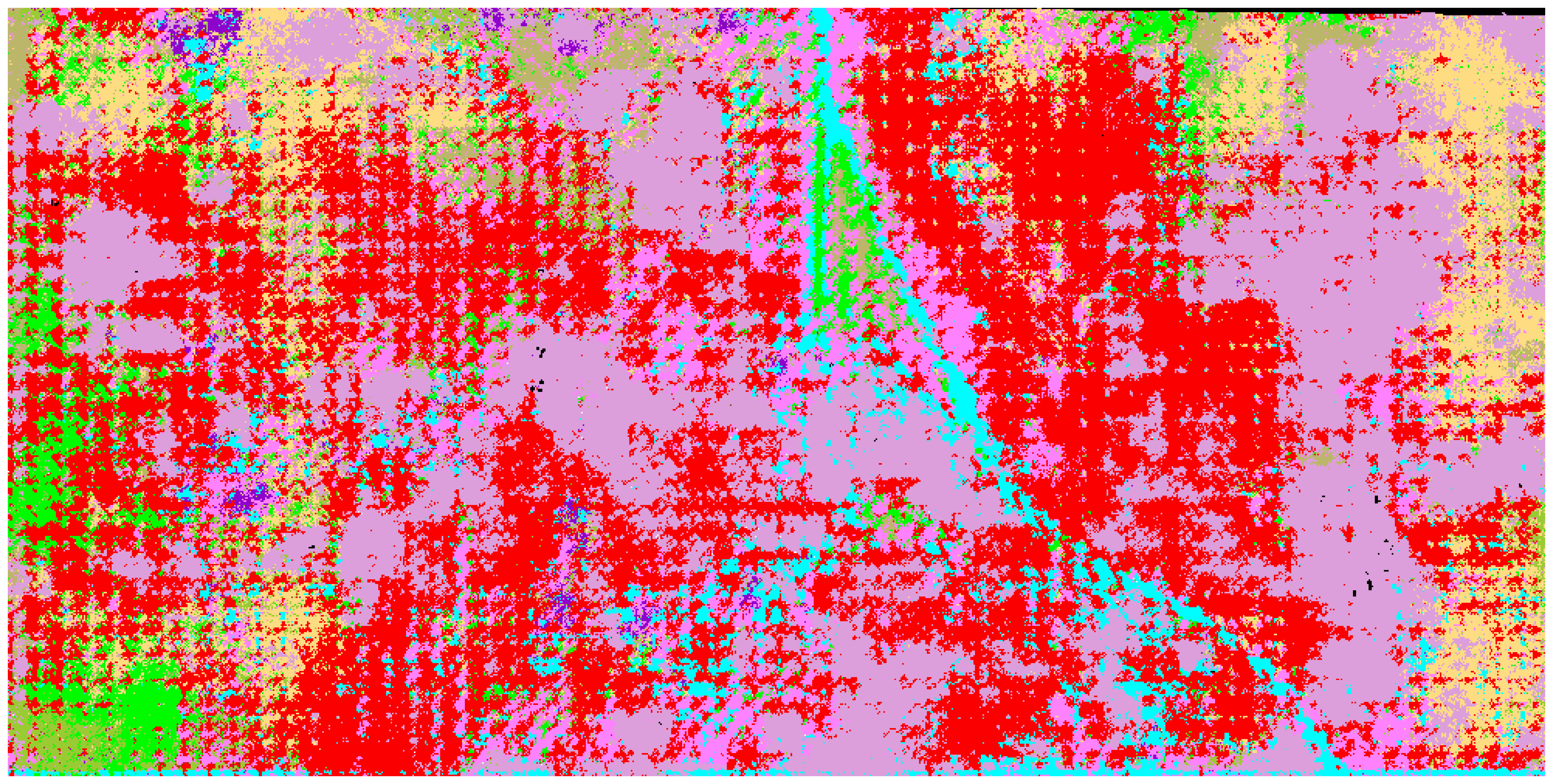}}

    \vspace{0.3cm} 
    \subfloat[]{\includegraphics[width=1\textwidth]{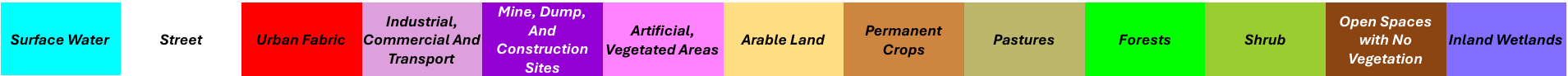}}

    \caption{Comparative Segmentation Inference using the C2Seg-AB Dataset. Display includes (a) Ground Truth Mask, (b) Our Method, (c) PCS, (d) Zero Shot, (e) GDA, (f) CDS,  (g) MIC, (h) CIA\_UDA, (i) UDA\_ME\_BS, and (j) Colorbar.}
    \label{fig:Germany_INFERENCE}
\end{figure*}

\cref{fig:Germany_INFERENCE} displays segmentation inference maps for this dataset. Consistent with the quantitative results in \cref{tab:Germany}, our proposed method performs comparably to PCS, with both showing superior performance relative to other methods. Notably, our method demonstrates improved segmentation for classes like Surface Water and Mine, Dump, and Construction Sites. It is worth mentioning that certain classes, such as Street, experience low precision across all methods, including ours.

\subsection{FLAIR}
\begin{figure*}[htbp]
    \centering
    \subfloat[]{\includegraphics[width=0.3\textwidth]{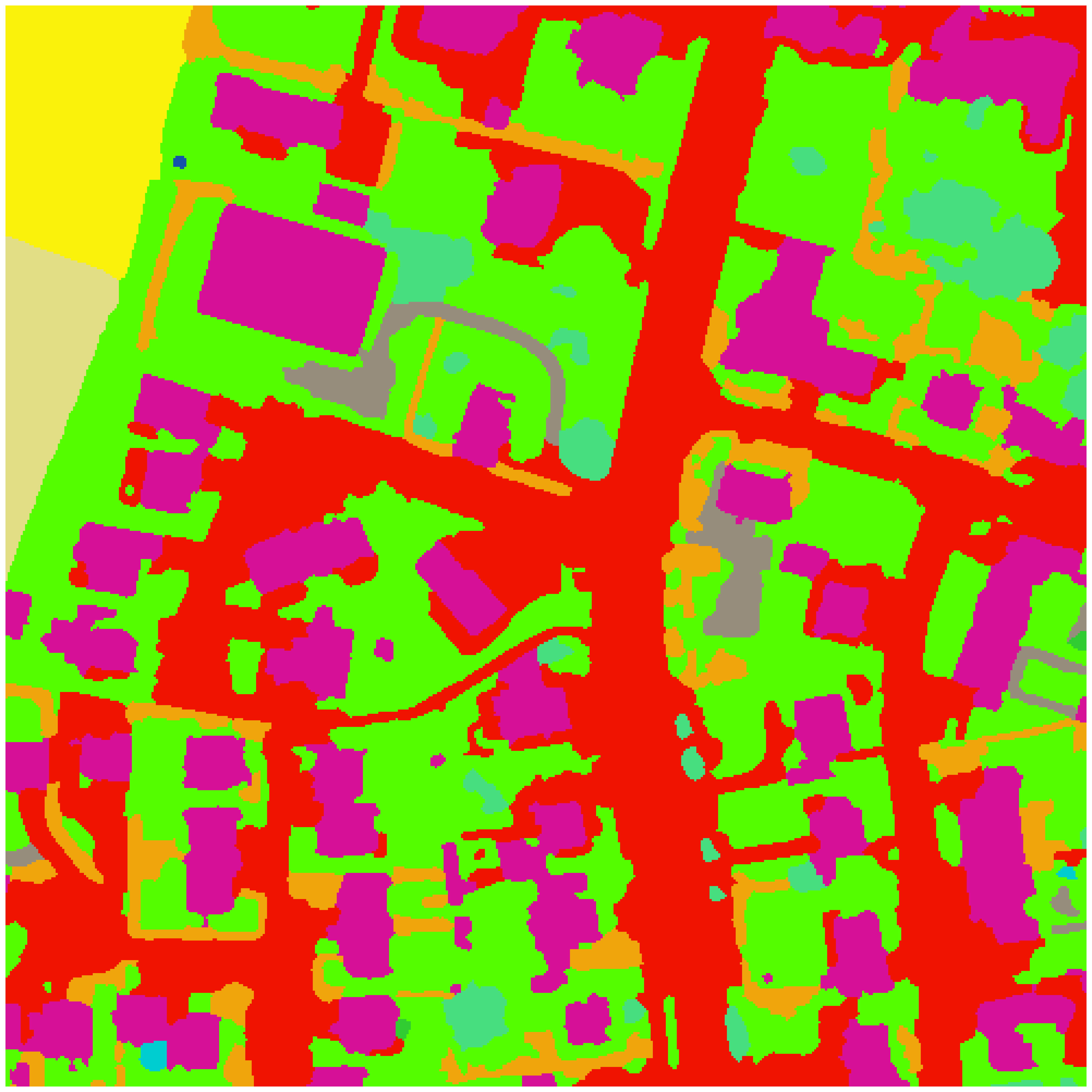}}
    \hfill
    \subfloat[]{\includegraphics[width=0.3\textwidth]{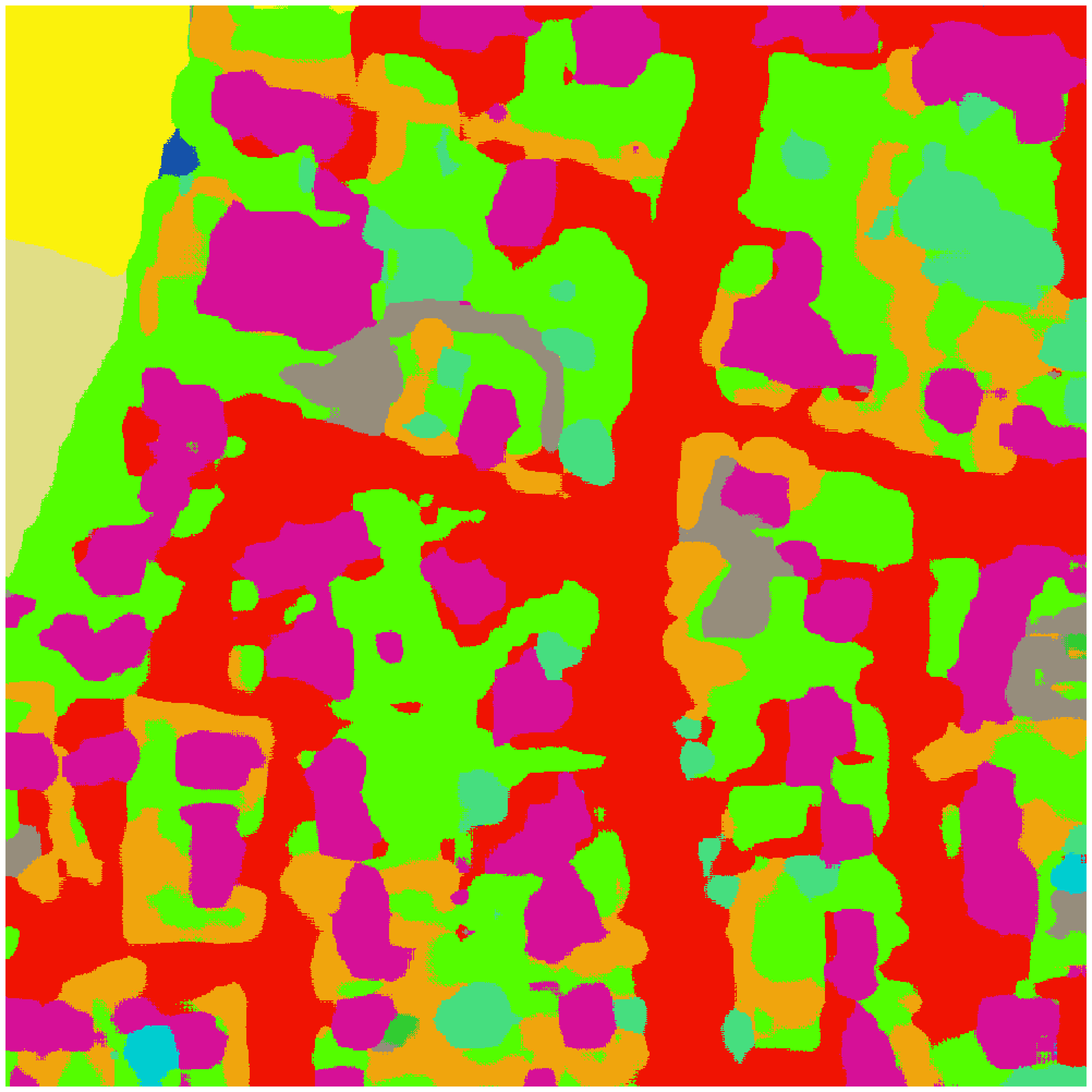}\label{subfig:2}}
    \hfill
    \subfloat[]{\includegraphics[width=0.3\textwidth]{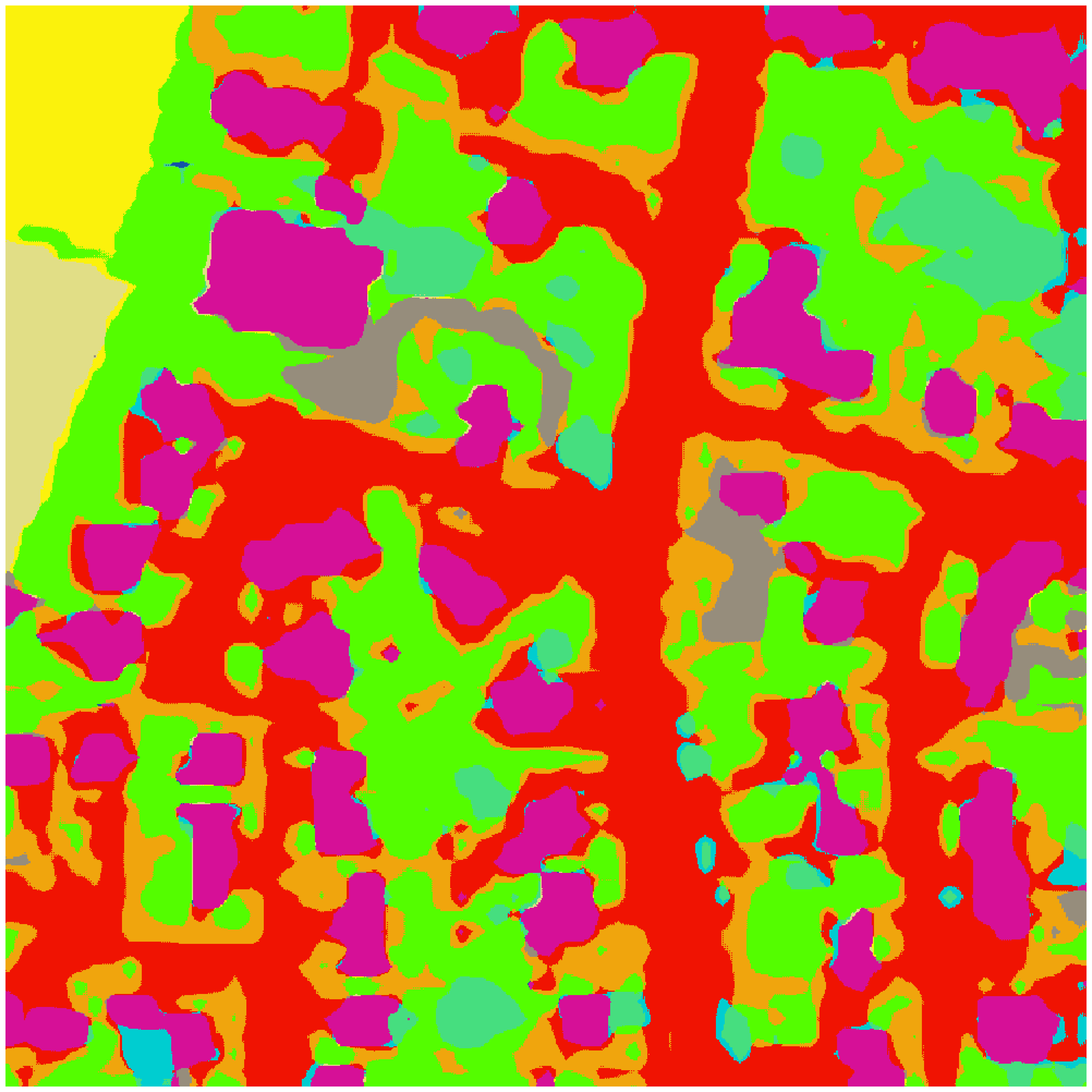}}
    
    \vspace{0.3cm} 

    \subfloat[]{\includegraphics[width=0.3\textwidth]{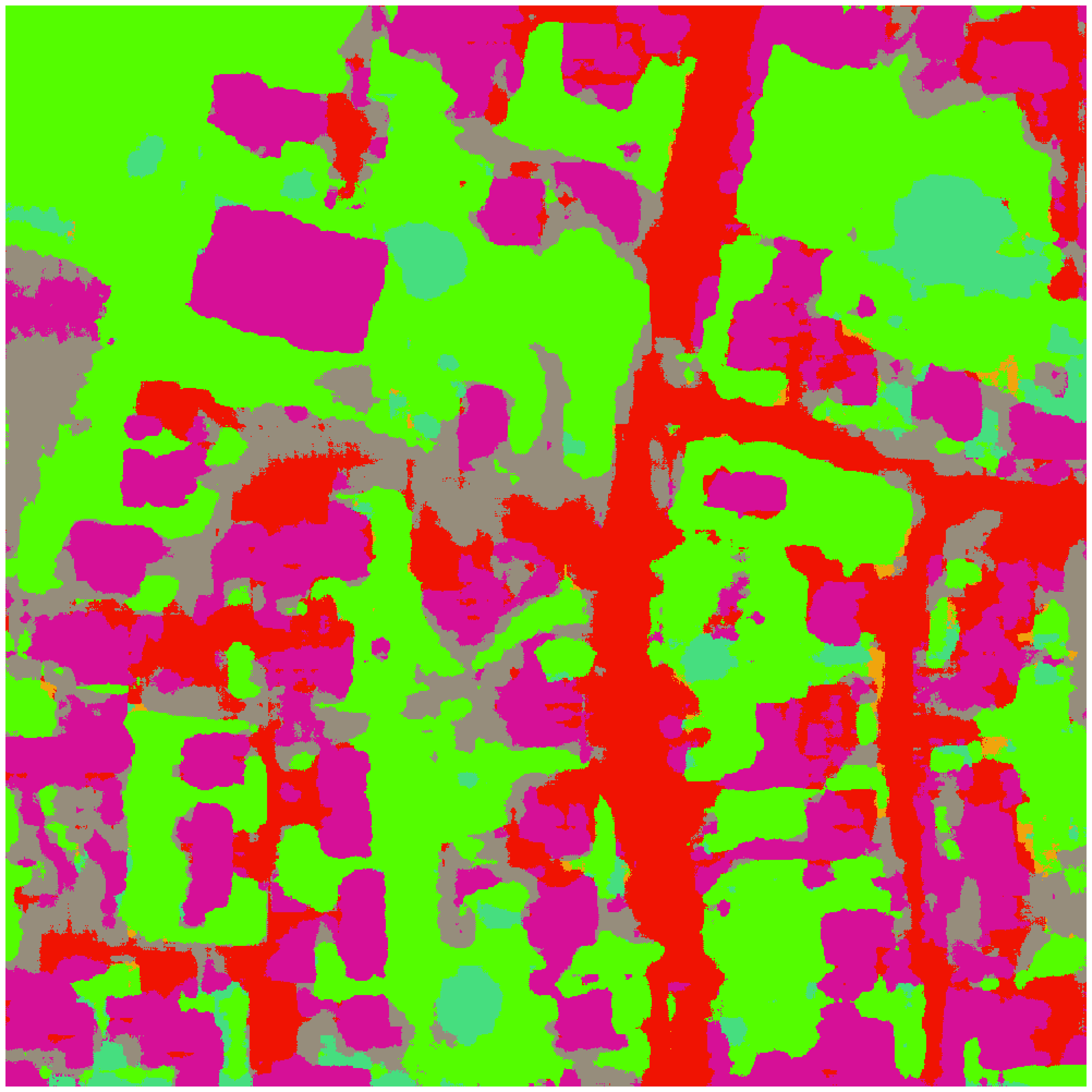}}
    \hfill
    \subfloat[]{\includegraphics[width=0.3\textwidth]{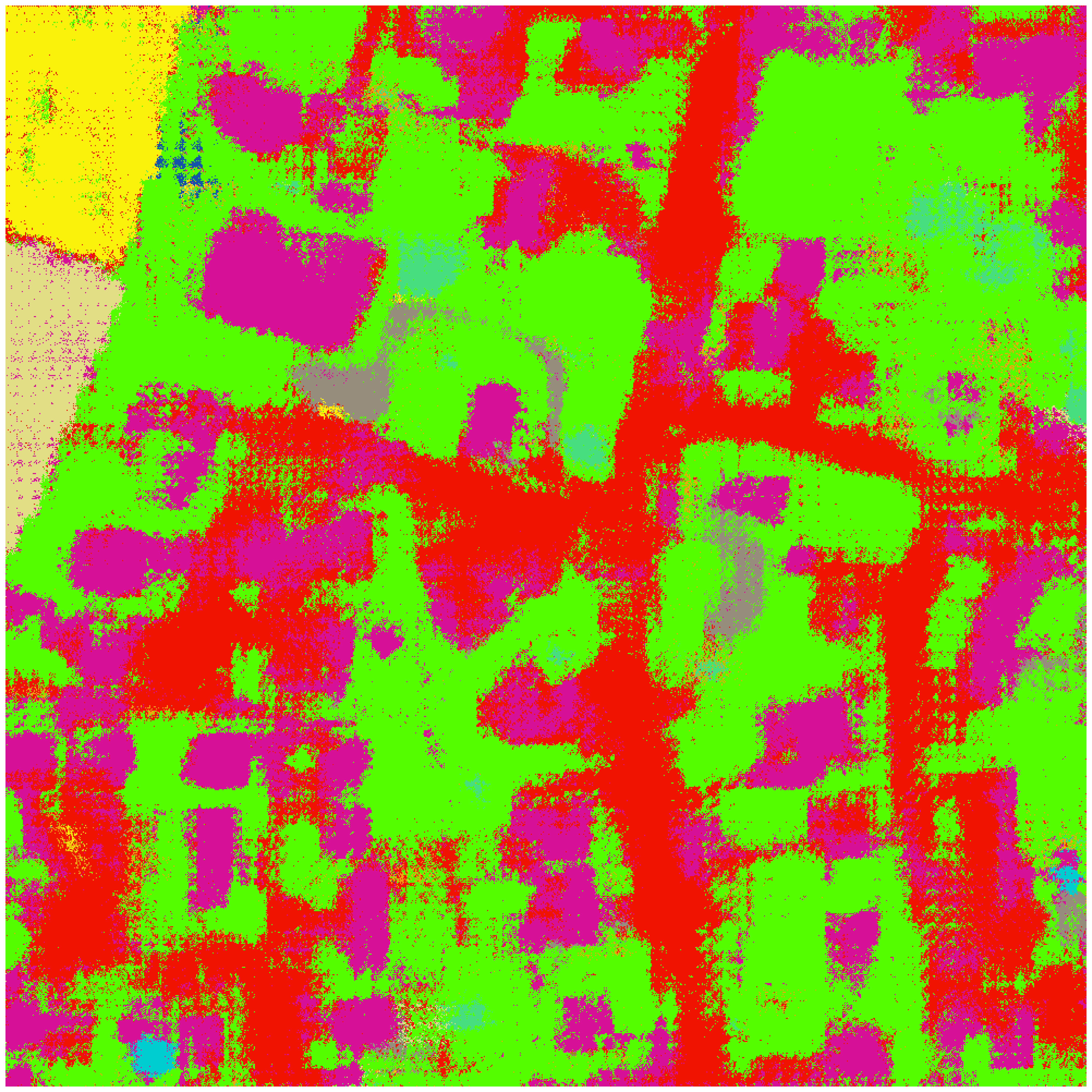}}
    \hfill
    \subfloat[]{\includegraphics[width=0.3\textwidth]{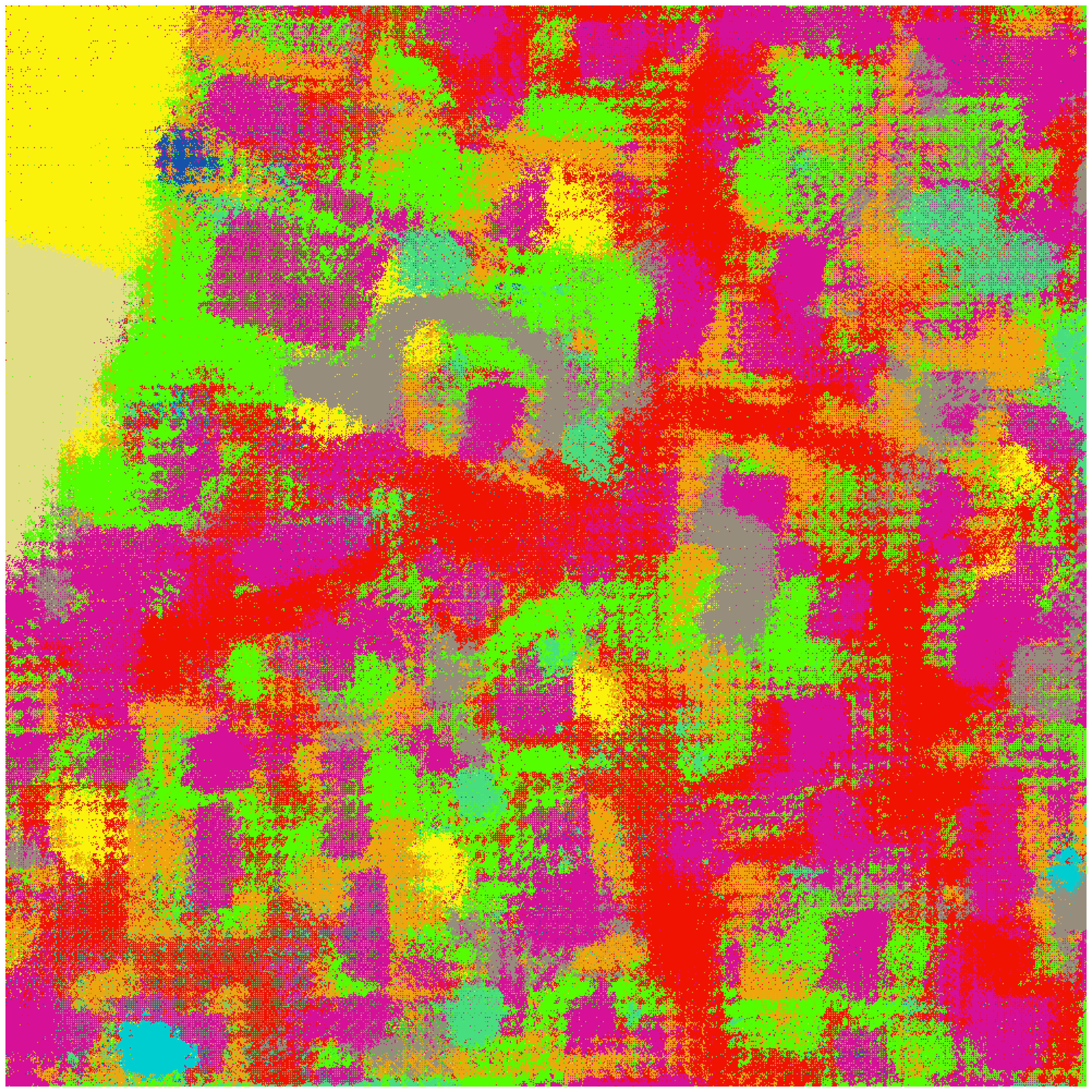}}
    
    \vspace{0.3cm} 

    \subfloat[]{\includegraphics[width=0.3\textwidth]{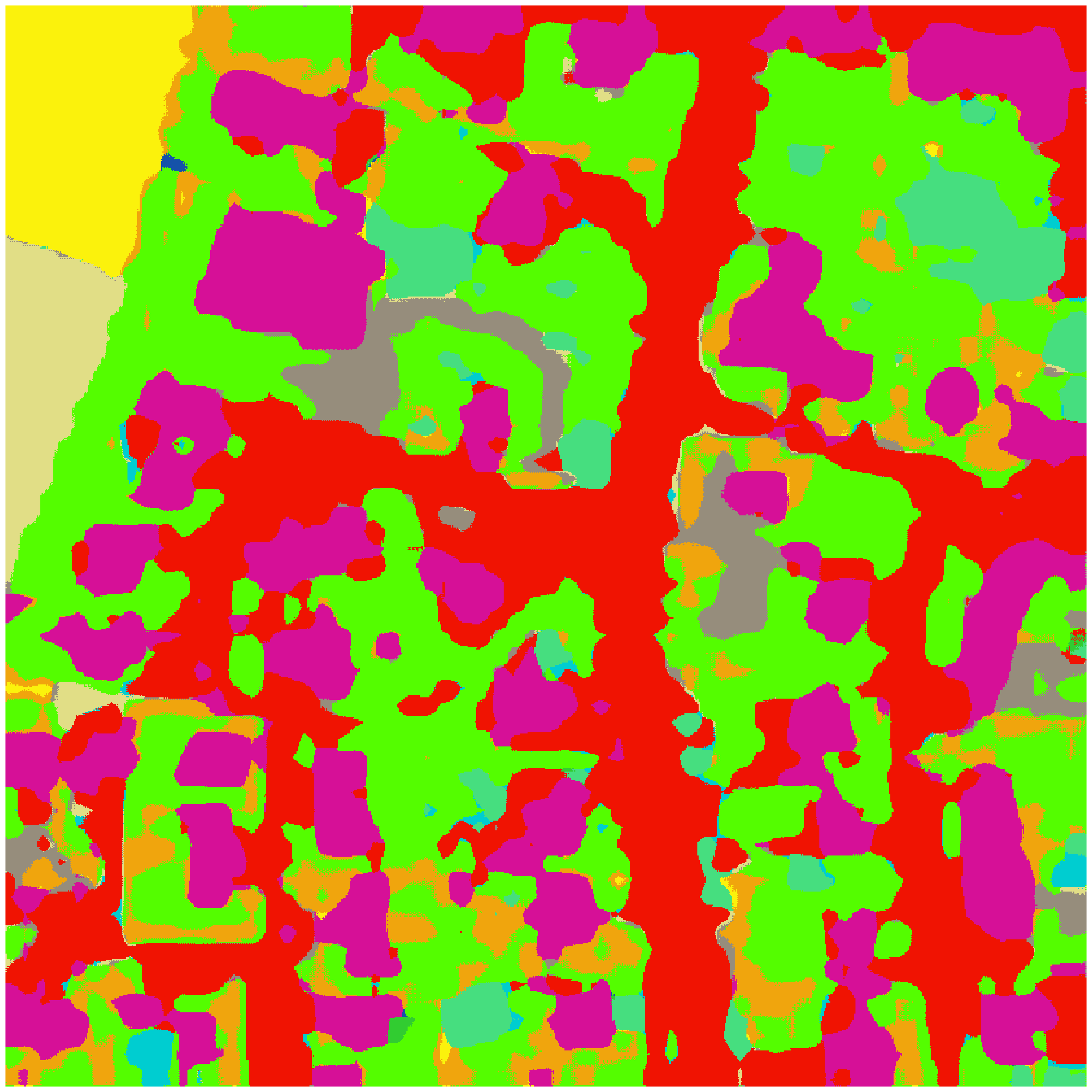}}
    \hfill
    \subfloat[]{\includegraphics[width=0.3\textwidth]{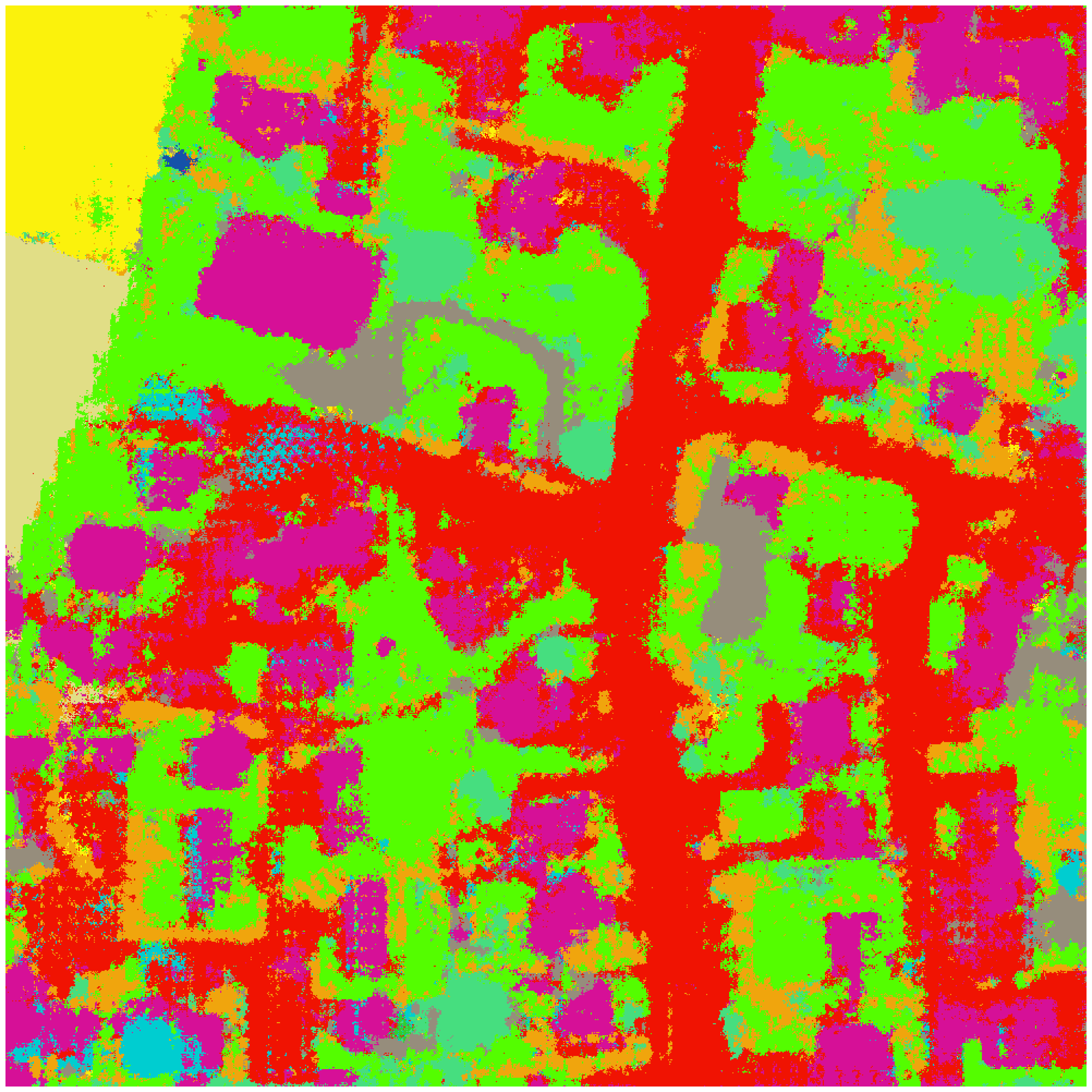}}
    \hfill
    \subfloat[]{\includegraphics[width=0.3\textwidth]{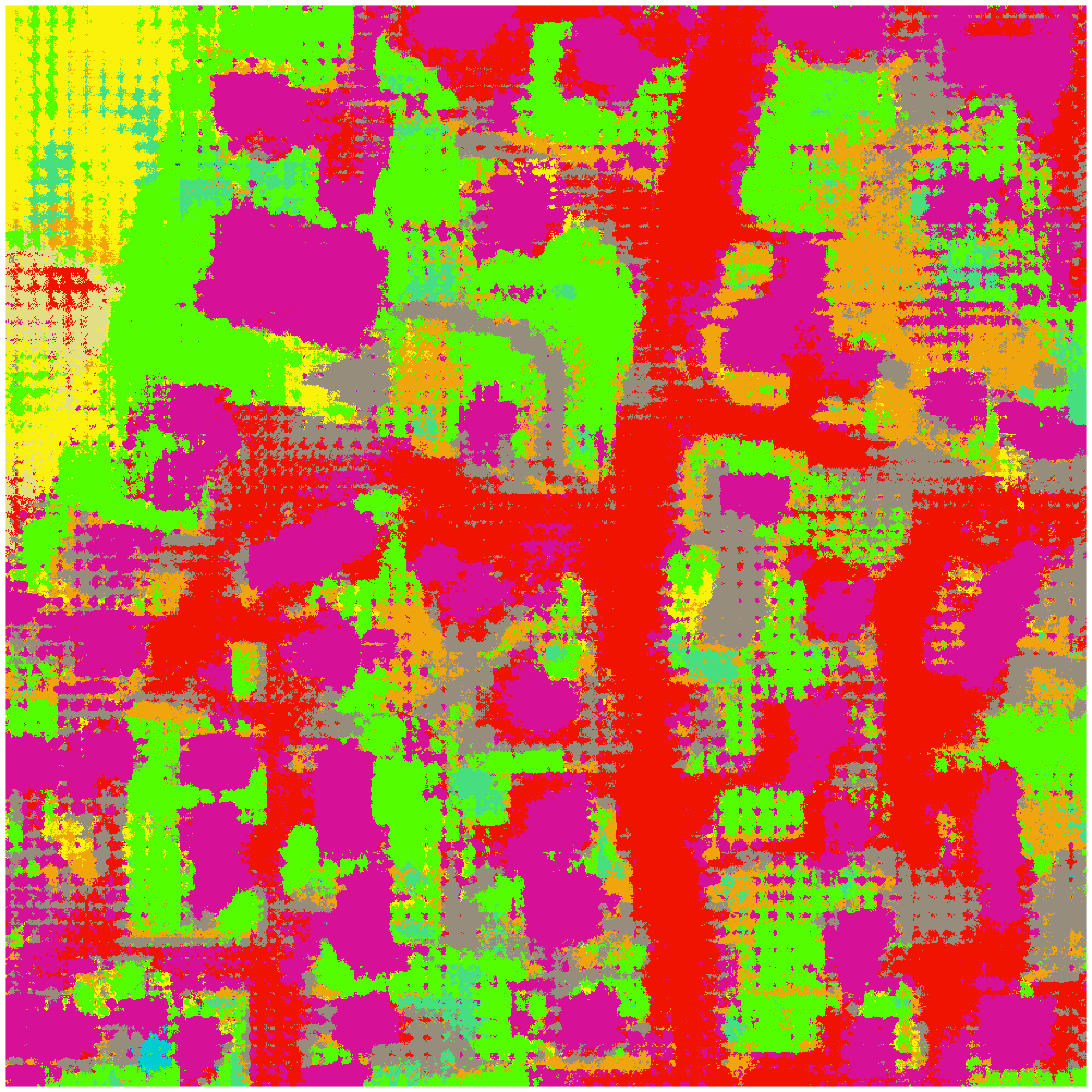}}

      \vspace{0.3cm}
       \subfloat[]{\includegraphics[width=1\textwidth]{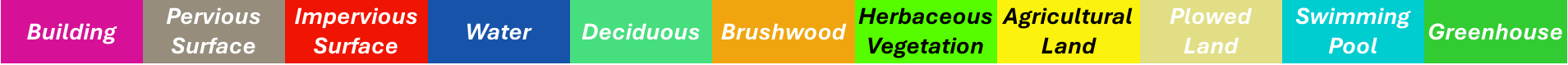}}
    \caption{Comparative Segmentation Inference using  FLAIR Dataset. Display includes (a) Ground Truth Mask, (b) Our Method, (c) PCS, (d) Zero Shot, (e) GDA, (f) CDS,  (g) MIC, (h) CIA\_UDA, (i) UDA\_ME\_BS, and (j) Colorbar.}
    \label{fig:FLAIR_INFERENCE}
\end{figure*}

\cref{fig:FLAIR_INFERENCE} displays segmentation inference maps for this dataset, highlighting that our approach achieves the most accurate segmentation map compared to other methods. Notably, our method demonstrates improved precision and recall for classes such as Agricultural Land, Plowed Land, and Brushwood.

\section{MAE-Based Generative Performance}
To evaluate the generative performance of our proposed model across different data modalities (MSI and HSI), we train it exclusively on the generative task represented primarily by $\mathcal{L}_{MAE}$ until convergence.{Specifically, this assessment aims to measure the model's ability to reconstruct the target domain image sequence from the concatenated source-target sequence, thereby encouraging the learning of generalizable domain-invariant features. For this evaluation, we employ a masking ratio of 50\% for target domain input images while keeping source domain images fully unmasked to generate the source-target sequence.}

\subsection{HSI}

\begin{figure*}[htbp]
    \centering
    \includegraphics[width=.99\textwidth]{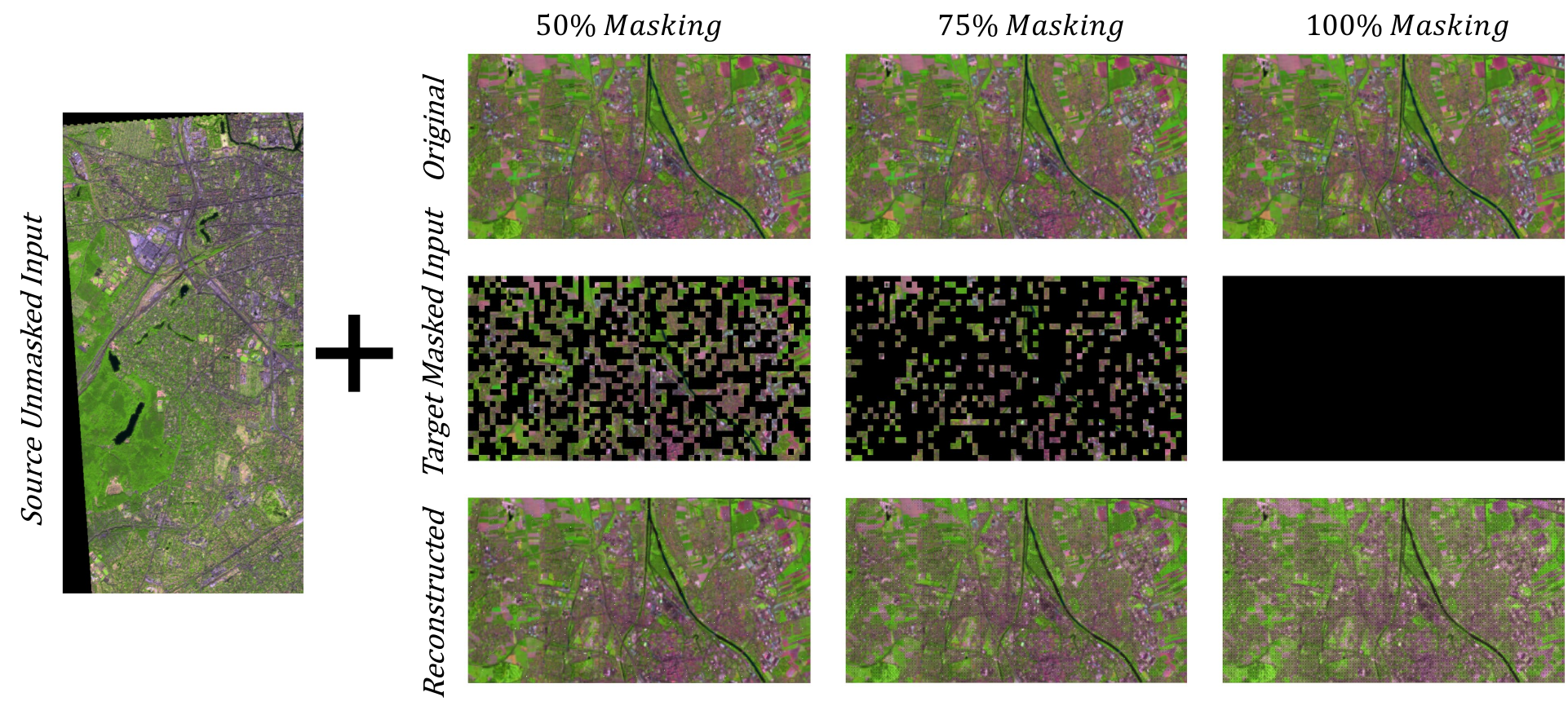}
    \caption{Generative task evaluation on HSI modality with C2Seg-AB dataset, showing original, masked, and reconstructed images across  three target domain masking levels.}
    \label{fig:Germany_rec}
\end{figure*}

\begin{figure*}
    \centering
    \begin{subfigure}[t]{\textwidth}
        \centering
        \includegraphics[width=\textwidth]{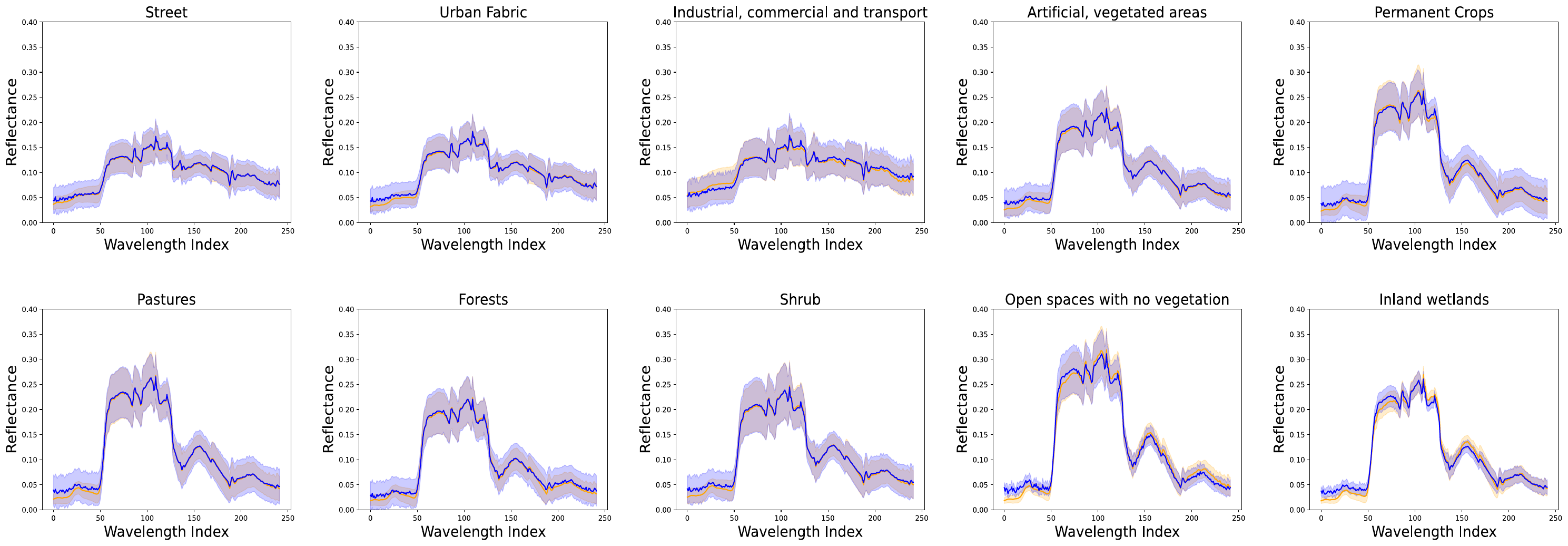}
        \caption{50\% Target Domain Masking}
    \end{subfigure}
    \vspace{5mm} 

    \begin{subfigure}[t]{\textwidth}
        \centering
        \includegraphics[width=\textwidth]{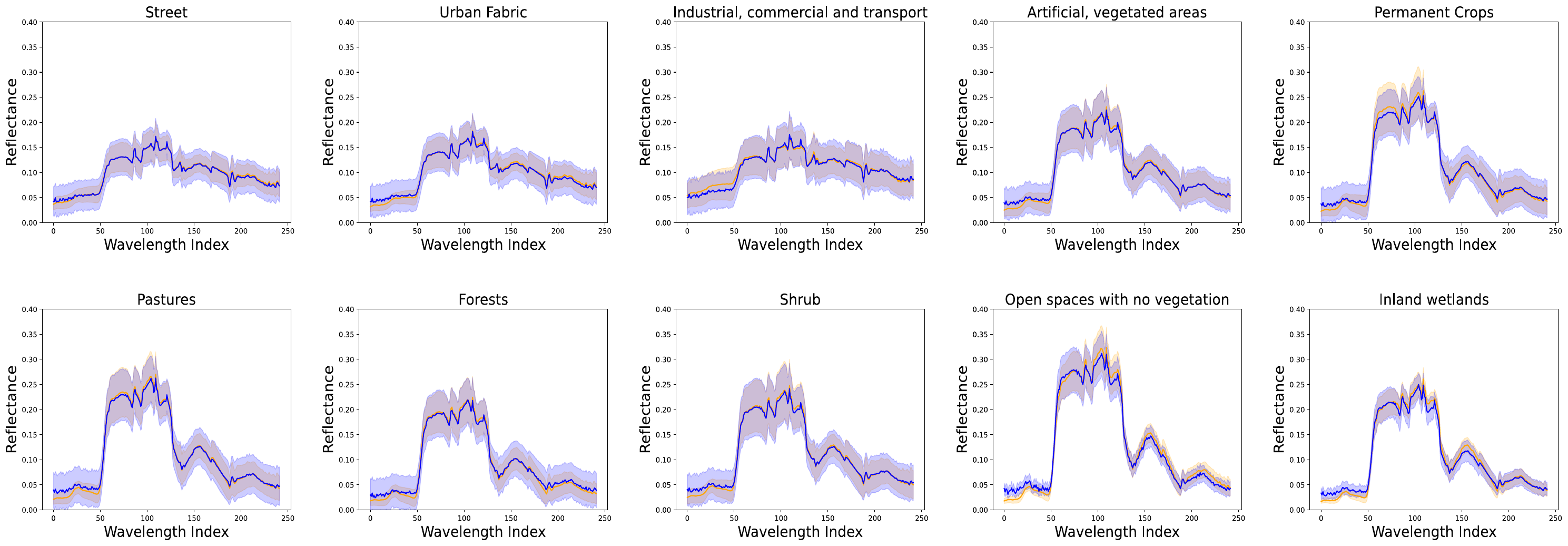} 
        \caption{75\% Target Domain Masking}
    \end{subfigure}
    \vspace{5mm} 

    \begin{subfigure}[t]{\textwidth}
        \centering
        \includegraphics[width=\textwidth]{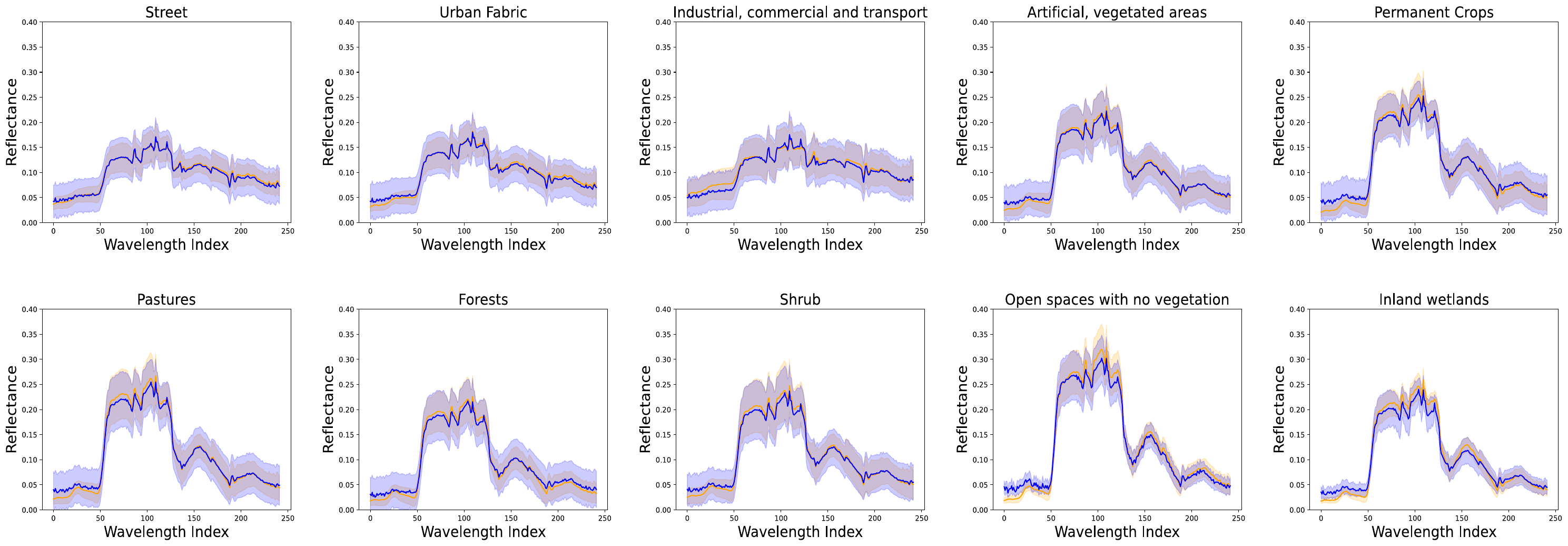} 
        \caption{100\% Target Domain Masking }
    \end{subfigure}

    \caption{Mean and standard deviation of the spectral distribution for the remaining classes not covered in the main text, evaluated for masked pixel classes in the target image at three different masking ratios using our MAE learning approach.}
    \label{fig:Rest_classes}
\end{figure*}

 { We evaluated the trained model on the HSI modality using the C2Seg-AB dataset, varying the masking ratio of the target domain images across 50\%, 75\%, and 100\%, while keeping the source domain images fully unmasked (0\% masking) across all experiments.} { To evaluate the reconstruction performance both spatially and spectrally, \cref{fig:Germany_rec} provides a visual assessment of spatial reconstruction quality, while \cref{fig:Rest_classes}  assess the reconstruction performance from a spectral perspective. The results from both spatial and spectral evaluations demonstrate strong reconstruction performance, highlighting the effectiveness of the proposed method.}
 This outcome highlights the model’s ability to capture informative features from one domain to aid reconstruction in another, promoting the learning of domain-invariant features. Importantly, our proposed framework extends MAE-based generative learning specifically to RS foundation models across multiple modalities, including HSI, which, as previously noted, has seen limited application with MAE-based self-learning approaches. For additional visual evaluation please refer to the supplementary material.

\subsection{MSI}
\begin{figure*}[htbp]
    \centering
    \includegraphics[width=.99\textwidth]{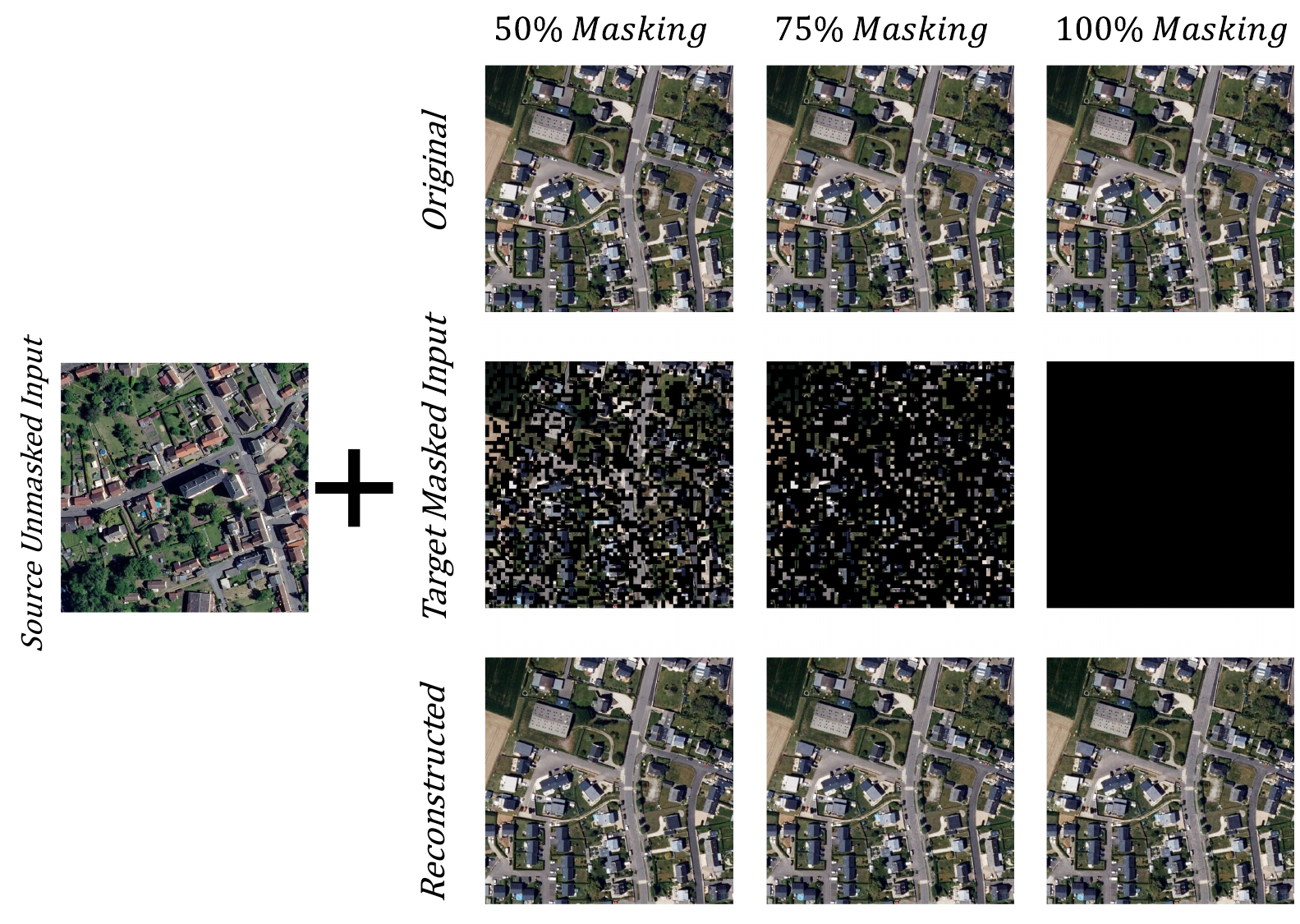}
    \caption{Generative task evaluation on MSI modality with FLAIR dataset, showing original, masked, and reconstructed images across three target domain masking levels.}
    \label{fig:flair_reconstruction}
\end{figure*}

 In a similar evaluation, we assessed the trained model on the MSI modality using the FLAIR dataset across various masking percentages for the target image, as shown in \cref{fig:flair_reconstruction}. The model consistently achieves near-perfect reconstruction quality under different masking ratios, demonstrating the flexibility and adaptability of our proposed method across multiple data modalities.

 \section{Additional Ablation Study Results}

\begin{table*}[htbp]
  \centering
  \caption{Contribution of each component in our proposed frame-
work to overall performance on the C2Seg-AB dataset.}
    
    \resizebox{.7\textwidth}{!}{
    \begin{tabular}{lrrrr}
    \toprule
    \multicolumn{1}{c}{\textbf{Classes}} & \multicolumn{1}{c}{$\mathcal{L}_{Seg}$} & \multicolumn{1}{c}{$\mathcal{L}_{Seg}+\mathcal{L}_{DA}$} & \multicolumn{1}{c}{$\mathcal{L}_{Seg}+\mathcal{L}_{MAE}$} & \multicolumn{1}{c}{$\mathcal{L}_{Seg}+\mathcal{L}_{DA}+\mathcal{L}_{MAE}$ (Ours)} \\
    \midrule
    Surface water & 0.4927 & 0.4971 & 0.4718 & 0.5138 \\
    Street & 0.2400 & 0.3021 & 0.1360 & 0.3207 \\
    Urban Fabric & 0.4555 & 0.6464 & 0.4950 & 0.6476 \\
    Industrial, commercial and transport & 0.5101 & 0.7295 & 0.5256 & 0.7376 \\
    Mine, dump, and construction sites & 0.5049 & 0.5611 & 0.3623 & 0.5949 \\
    Artificial, vegetated areas & 0.4314 & 0.6504 & 0.4136 & 0.6615 \\
    Arable Land & 0.4894 & 0.8324 & 0.4725 & 0.8382 \\
    Permanent Crops & 0.2434 & 0.2569 & 0.1130 & 0.2358 \\
    Pastures & 0.5354 & 0.6311 & 0.4016 & 0.6451 \\
    Forests & 0.5686 & 0.6144 & 0.5022 & 0.6247 \\
    Shrub & 0.3960 & 0.5305 & 0.2924 & 0.5404 \\
    Open spaces with no vegetation & 0.0956 & 0.0140 & 0.0231 & 0.0217 \\
    Inland wetlands & 0.3783 & 0.4423 & 0.2665 & 0.4503 \\
     \midrule
    MA(Avg) & 0.4950 & 0.6252 & 0.3930 & 0.6381 \\
    
    MA(Std) & 0.0621 & 0.0220 & 0.0654 & 0.0208 \\
    mIoU (Avg) & 0.2704 & 0.3741 & 0.2245 & 0.3835 \\
    mIoU (Std) & 0.0327 & 0.0174 & 0.0331 & 0.0161 \\
    mF1 (Avg) & 0.4109 & 0.5160 & 0.3443 & 0.5255 \\
    mF1 (Std) & 0.0420 & 0.0187 & 0.0482 & 0.0186 \\
    \end{tabular}%
    }
  \label{tab:German_Abiliation}%
\end{table*}%

Similar to \cref{tab:Abilation_Flair} in terms of the baseline represented by the second column, we observe that adding $\mathcal{L}_{MAE}$ alone degrades performance compared to the baseline, which can be attributed to the dynamic weighting mechanism discussed in the mathematical insight section. This mechanism depends on the model's predictions for unlabeled samples in the target domain, leading to potential issues when the model is mistakenly confident about these samples. However, combining $\mathcal{L}_{MAE}$ with $\mathcal{L}_{DA}$ yields the best performance, consistent with the previously described synergy between these two components.

\end{document}